\documentclass[letterpaper,journal]{IEEEtran}
\usepackage{amsmath,amsfonts}
\usepackage{algorithmic}
\usepackage{algorithm}
\usepackage{array}
\usepackage[caption=false,font=normalsize,labelfont=sf,textfont=sf]{subfig}
\usepackage{textcomp}
\usepackage{stfloats}
\usepackage{url}
\usepackage{verbatim}
\usepackage{graphicx}
\usepackage{cite}
\newcommand{\etal}{\textit{et al.}}
\newcommand{\ie}{\textit{i.e.}}
\newcommand{\eg}{\textit{e.g.}}

\usepackage{gensymb}
\usepackage{multirow}
\usepackage{multicol}
\usepackage{booktabs}
\usepackage{color,soul}

\begin{document}

\title{A Generic Hybrid Framework \\
for 2D Visual Reconstruction}

\author{Daniel Rika$^{a}$, Dror Sholomon$^{a}$, Eli O. David$^{a}$, Alexandre Pais$^{b}$, and Nathan S. Netanyahu$^{a,c}$%
\thanks{$^{a}$ Department of Computer Science, Bar-Ilan University, Ramat Gan 5290002, Israel. Contact emails: danielrika@gmail.com, dror.sholomon@gmail.com, dr.elidavid@gmail.com, nathan@cs.biu.ac.il.}%
\thanks{$^{b}$ National Tile Museum, Lisbon 1900-312, Portugal. Contact email: apais@mnazulejo.dgpc.pt.}%
\thanks{$^{c}$ Data Science and AI Institute, Bar-Ilan University, Ramat Gan 5290002, Israel.}
}

\maketitle

\begin{abstract}
This paper presents a versatile hybrid framework for addressing 2D real-world reconstruction tasks formulated as jigsaw puzzle problems (JPPs) with square, non-overlapping pieces. Our approach integrates a deep learning (DL)-based compatibility measure (CM) model that evaluates pairs of puzzle pieces holistically, rather than focusing solely on their adjacent edges as traditionally done. This DL-based CM is paired with an optimized genetic algorithm (GA)-based solver, which iteratively searches for a global optimal arrangement using the pairwise CM scores of the puzzle pieces. Extensive experimental results highlight the framework's adaptability and robustness across multiple real-world domains. Notably, our unique hybrid methodology achieves state-of-the-art (SOTA) results in reconstructing Portuguese tile panels and large degraded puzzles with eroded boundaries.
\end{abstract}


\section{Introduction}
\label{sec:introduction}
\IEEEPARstart{O}{bject} reconstruction from fragmented pieces is a fundamental problem with broad significance across diverse fields, including archaeology, art restoration, and document forensics. Examples range from reassembling broken pottery and ancient frescoes to reconstructing shredded documents. At its core, this task can be abstracted as assembling an object from $N$ distinct, non-overlapping pieces with the goal of achieving the most accurate and efficient reconstruction. Typically, these pieces are represented as colored image fragments, making the problem closely aligned with the well-known {\em jigsaw puzzle problem} (JPP), which is computationally classified as NP-complete~\cite{journals/aai/Altman89,springerlink:10.1007/s00373-007-0713-4}.

The JPP has been widely employed as a testbed for real-world challenges, including image editing~\cite{bb43059}, the recovery of shredded documents and photographs~\cite{liu2011automated,marques2009reconstructing,justino2006reconstructing,conf/icip/DeeverG12}, art conservation~\cite{journals/tog/BrownTNBDVDRW08,journals/KollerL06,andalo2016impa}, and speech descrambling~\cite{Zhao:2007:PSA:1348258.1348289,chuman2017}. However, real-world reconstruction tasks often deviate significantly from the idealized JPP setting. Practical challenges include missing pieces, gaps resulting from material degradation, unknown image dimensions, and the presence of multiple fragmented datasets.

Most practical reconstruction schemes require typically a {\em compatibility measure} that assesses the likelihood of adjacency between two given pieces, alongside an effective strategy for global piece placement. Traditional CMs, which rely on low-level color and texture correlations near tile boundaries, often fall short when applied to real-world problems. For example, archaeological fragments and shredded documents frequently exhibit severe degradation at their edges, while Portuguese tile panels often lack sufficient chromatic variability, making adjacency detection unreliable. Furthermore, existing solvers typically adopt greedy strategies, which struggle with local optima arising from the inaccuracies in such measures.

\begin{figure*}[!t]
    \centering
    \subfloat[Scrambled]{\includegraphics[width=0.4\linewidth]{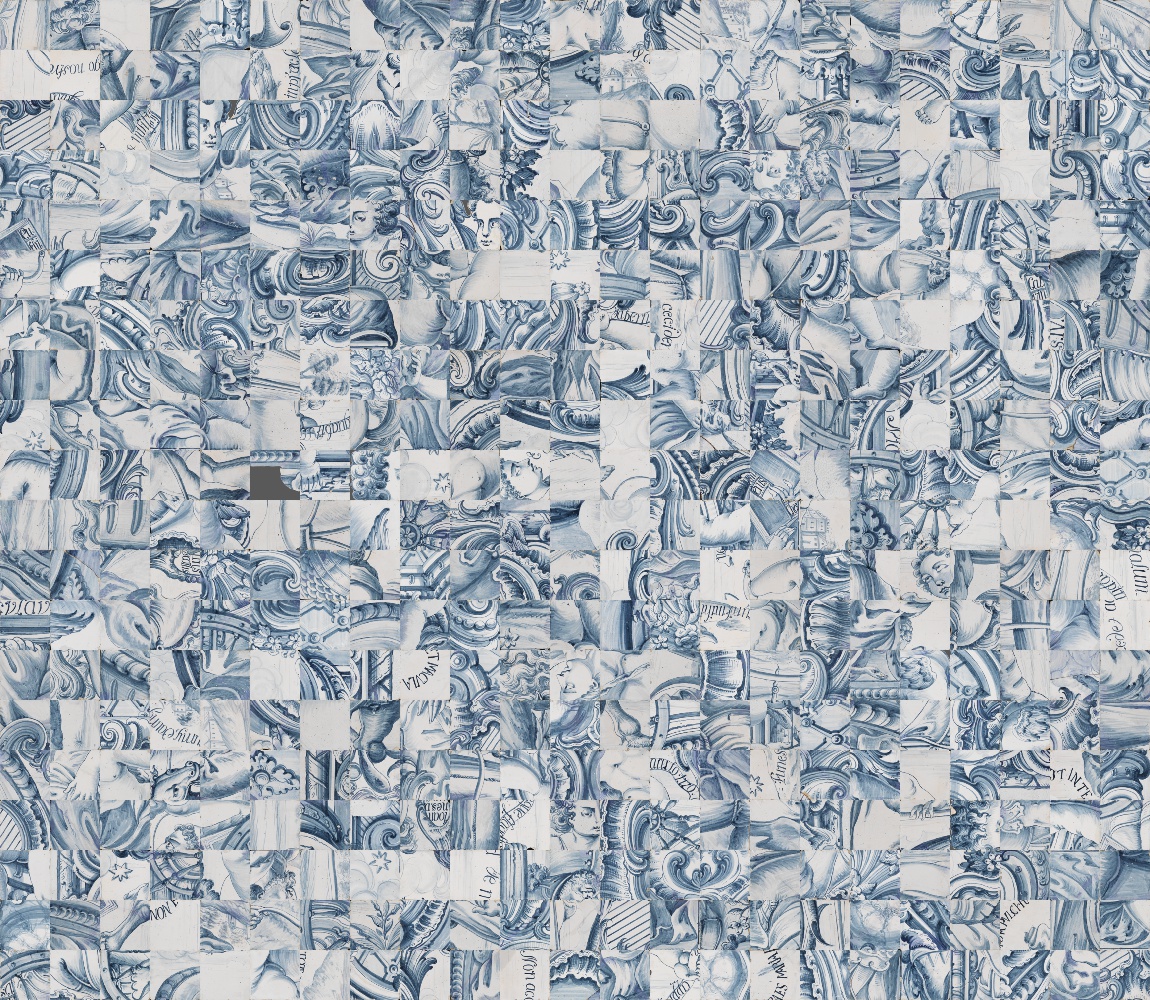}}
    \hfil
    \subfloat[Perfectly reconstructed]{\includegraphics[width=0.4\linewidth]{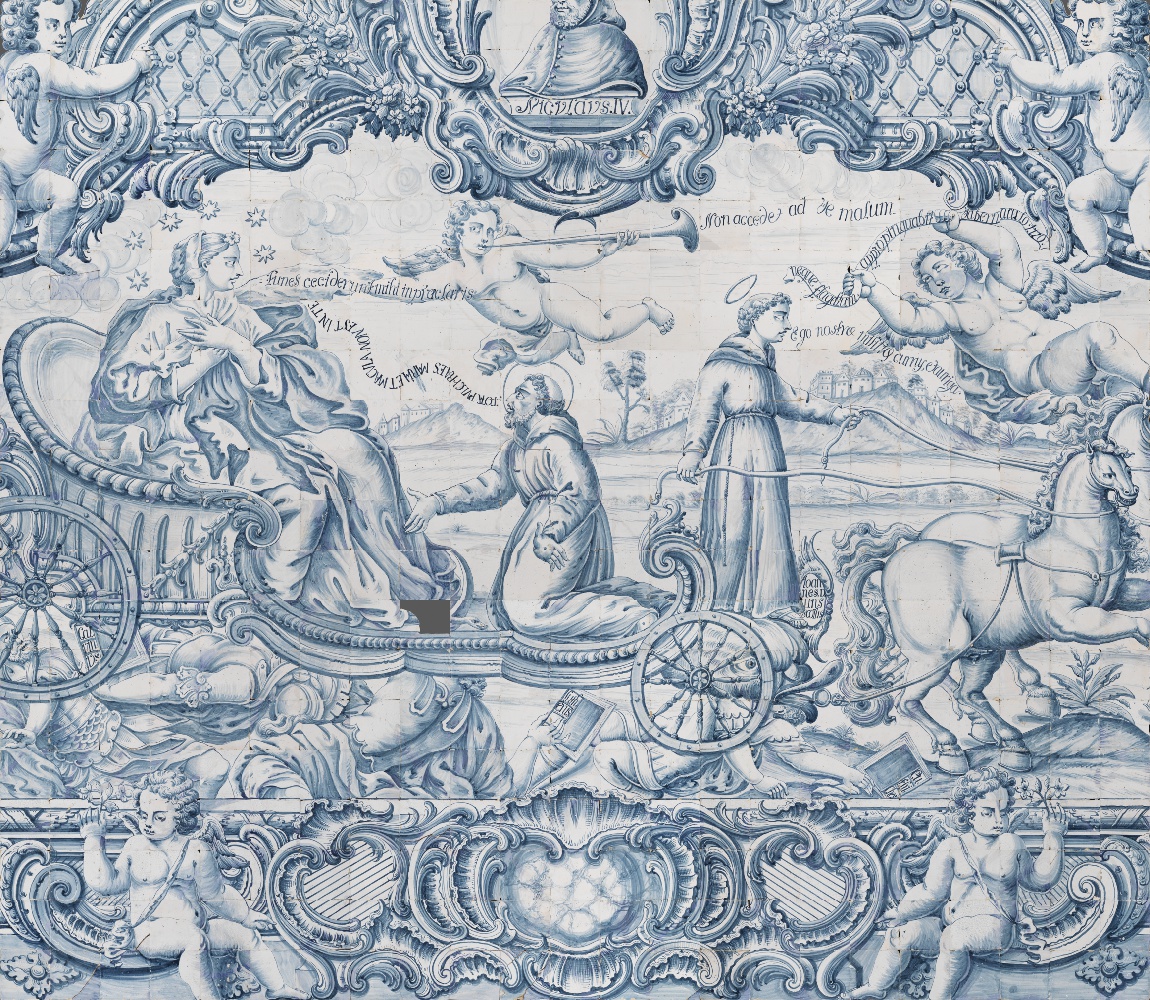}}
    \caption{Reconstruction of 460-tile Portuguese panel with unknown piece orientation and panel dimensions using our proposed system: \textbf{(a)} Scrambled image of 460-piece panel, and \textbf{(b)} perfectly reconstructed panel achieved through our deep-learning compatibility measure (DLCM) and GA-based solver.}
    \label{fig:reconstruction_front_page}
\end{figure*}

To address these limitations, we propose a generic {\em computational intelligence} (CI) framework that synergizes {\em deep learning} (DL) and {\em evolutionary computation} (EC) techniques~\cite{Sholomon_2013_CVPR,sholomon2014genetic,rika2019gecco}. Specifically, our framework consists of: (1) An innovative DL-based compatibility measure (CM) model, which evaluates adjacency by analyzing entire puzzle pieces rather than focusing solely on their boundaries. This approach eliminates the need for hand-crafted feature extraction and improves robustness across varying datasets, and (2) An enhanced genetic algorithm (GA)-based solver, which iteratively optimizes piece placement by leveraging biologically-inspired evolutionary strategies, effectively overcoming local optima and achieving near-global reconstruction accuracy. Figure~\ref{fig:reconstruction_front_page} showcases a successful reconstruction of a complex 460-piece Portuguese tile panel using our proposed methodology.

The key contributions of our work are the following:
\begin{enumerate}
\item
We introduce a hybrid framework combining a DL-based compatibility measure with an enhanced GA-based solver for robust 2D visual reconstruction.
\item
We demonstrate our framework's applicability across diverse real-world scenarios, including Portuguese tile panels, degraded puzzles with eroded boundaries, and shredded documents, validated through extensive empirical analysis.
\item
We achieve state-of-the-art (SOTA) performance across multiple benchmarks, outperforming existing methods in most evaluated domains.
\item
We curate and release a new benchmark dataset of Portuguese tile panels to support future research in this field.
\end{enumerate}

This novel hybrid approach establishes a scalable and adaptable methodology for addressing complex 2D reconstruction challenges, offering both theoretical insights and practical advancements for real-world applications.

The remainder of this paper is structured as follows: Section~\ref{sec:related_work} reviews prior research on jigsaw puzzle problem (JPP) methodologies, including generative deep learning (DL)-based compatibility measures (CMs), end-to-end (E2E) frameworks, and recent approaches addressing real-world reconstruction challenges. Section~\ref{sec:cnn_based_cm} details our specialized DL-based CM model and its training process. Section~\ref{sec:ga_solver} describes our enhanced GA-based reconstruction solver, which complements the proposed CM model. Extensive experimental results are presented in Sections~\ref{sec:experimental_results} and \ref{sec:other_problem_domains}, covering the Portuguese tile problem, synthetic JPPs, and additional real-world domains, including puzzles with eroded boundaries and shredded documents. Finally, Section~\ref{sec:conclusions} concludes the paper with key insights and future directions.

\section{Related Work}
\label{sec:related_work}
\subsection{Main Methodologies}
\subsubsection{\textbf{Traditional Methods}}
Early work on the JPP can be traced back to Freeman and Garder~\cite{bb47278}, who, in 1964, introduced a computational solver capable of handling puzzles with up to nine pieces. Subsequent studies~\cite{radack1982jigsaw,wolfson1988solving,kong2001solving,goldberg2002global} focused exclusively on the shape features of puzzle pieces.

Kosiba {\etal}~\cite{kosiba1994automatic} were the first to incorporate image content alongside boundary shape. Their method calculated color compatibility along matching contours, rewarding adjacent pieces with similar colors. This approach dominated research for over a decade (see, {\eg}, \cite{chung1998jigsaw,yao2003shape,makridis2006new,sagiroglu2006texture,nielsen2008solving}) until the focus shifted toward color-based solvers for square-tile puzzles with predefined piece orientations ({\ie}, \textit{Type-1} puzzles).

Cho {\etal}~\cite{conf/cvpr/ChoAF10} employed {\em dissimilarity} ({\ie}, the sum of squared color differences across all neighboring pixels and color bands) as a compatibility measure (CM) in their probabilistic puzzle solver, which could handle puzzles with up to 432 pieces. Yang {\etal}~\cite{yang2011particle} enhanced this approach with a {\em particle filter}-based solver, achieving further improvements. Shortly thereafter, Pomeranz {\etal}~\cite{conf/cvpr/PomeranzSB11} introduced the first fully automated solver capable of reconstructing puzzles with up to 3,000 square pieces, leveraging dissimilarity and their innovative {\em best-buddies} heuristic.

Gallagher~\cite{gallagher2012jigsaw} significantly advanced the SOTA by tackling puzzles with unknown piece orientations ({\ie}, \textit{Type-2} puzzles) and puzzles with undefined dimensions. He introduced the {\em Mahalanobis gradient compatibility} (MGC) measure, which penalizes gradient intensity variations and incorporates the covariance of color channels to enhance compatibility. Additionally, he proposed using {\em dissimilarity ratios} to provide a more robust compatibility measure. Gallagher's reconstruction approach drew inspiration from Kruskal's minimum spanning tree algorithm, merging subtrees based on compatibility measures (CMs) while adhering to geometric constraints.  

Sholomon {\etal}~\cite{Sholomon_2013_CVPR,sholomon2014genetic,sholomon2014generalized} adopted a genetic algorithm (GA)-based approach featuring innovative {\em crossover} procedures. Their methodology demonstrated strong performance on large-scale Type-1 and Type-2 puzzles, including two-sided and mixed puzzles.  

Son {\etal}~\cite{son2014solving} introduced {\em loop constraints}, where the dissimilarity ratio for each consecutive pair of pieces in loops of four or more must remain below a specified threshold relative to the smallest edge distance. This approach improved accuracy for certain Type-1 and Type-2 puzzles. They also established, for the first time, upper bounds on reconstruction accuracy across various datasets. In subsequent studies~\cite{son2016cvpr, son2019tpami}, they proposed a novel assembly strategy based on achieving {\em maximum geometric consensus} among pieces using {\em hierarchical piece loops}.  

Yu {\etal}~\cite{yu2016bmvc} reformulated the jigsaw puzzle problem (JPP) as a {\em linear programming} task to compute global position scores for each piece, enhancing robustness through the use of a weighted $L_1$ penalty.  

Huroyan~{\etal}~\cite{laplacia2020siam} addressed Type-2 puzzles by first determining piece orientations using a \textit{graph connection Laplacian} (GCL) and subsequently solving the puzzle using the Type-1 solver proposed by Yu {\etal}~\cite{yu2016bmvc}.  

Paikin and Tal~\cite{paikin2015solving} proposed a greedy solver utilizing an asymmetric $L_1$-norm dissimilarity and the best-buddies heuristic. Their approach effectively handled puzzles with missing pieces, improving both accuracy and runtime. Later, Andal{\'o} {\etal}~\cite{7442162} mapped the JPP to a constrained quadratic optimization problem and introduced the deterministic \textit{PSQP algorithm}, which solves it via gradient ascent.  

\subsubsection{\textbf{DL-Based Pairwise Compatibilities}}
The {\em DNN-buddies} heuristic~\cite{sholomon2016dnn} was initially introduced to improve the accuracy of a GA-based solver proposed in~\cite{sholomon2014generalized}.

Pai\~{x}ao \etal~\cite{paixao2018deep} utilized SqueezeNet~\cite{SqueezeNet} and MobileNetV2~\cite{mobilenetv2_2018_cvpr} as deep learning (DL)-based compatibility measures (CMs) for reconstructing \textit{strip-cut} shredded documents. In subsequent work, they developed a more effective approach by incorporating piece edge embeddings~\cite{paixão2020faster}. Both methods were paired with a puzzle solver for the \textit{asymmetric traveling salesman problem} (ATSP).

Bridger \etal~\cite{bridger2020cvpr} introduced a compatibility measure derived from a \textit{generative adversarial network} (GAN)~\cite{gan2014} to solve puzzles with eroded boundaries. Their method employed the \textit{generator} to fill gaps between puzzle pieces, while the \textit{discriminator} assigned compatibility scores by learning to distinguish between plausible and implausible in-painting pairs. This GAN-based CM was integrated with the solver by Paikin and Tal~\cite{paikin2015solving} for puzzle reconstruction.

Khoroshiltseva \textit{et al.}~\cite{JiGAN2022iciap} also focused on reconstructing eroded puzzles with their \textit{JiGAN} framework. Their approach first used a GAN-based model to extend images beyond their original boundaries, followed by Gallagher's MGC~\cite{gallagher2012jigsaw} for pairwise compatibility estimation and a \textit{relaxation labeling} algorithm~\cite{relaxtionLabeling2021caip} for final reconstruction. Despite the improved computational efficiency, their results were inferior to those achieved by Bridger \etal~\cite{bridger2020cvpr}.

Song~\etal~\cite{10096300} proposed the \textit{per-fragment puzzlet discriminant network} (PF-PDN), which employed a convolutional deep learning (CDL)-based Siamese network to assess individual fragments, and the \textit{per-image puzzlet discriminant network} (PI-PDN), designed to evaluate entire puzzlets directly. These models were combined with a GA-based solver and demonstrated on puzzles of size ($3 \times 3$ and) $5 \times 5$ with eroded boundaries.

\subsubsection{\textbf{End-to-End Schemes}}
Recent end-to-end (E2E) approaches to the JPP have adopted single, unified deep learning (DL)-based frameworks rather than integrating a compatibility measure (CM) module with a separate solver.  

Doersch~\etal~\cite{doersch2015unsupervised}, Noroozi and Favaro~\cite{noroozi2016unsupervised}, Dery~\etal~\cite{dery2017jpp}, and Santa Cruze~\etal~\cite{santacruz2017visualpermutation} were among the first to explore this trend. Their primary objective was to ``repurpose'' neural networks trained to solve small-scale puzzles for more advanced tasks, such as object detection and classification, in an unsupervised manner.  

Paumard \textit{et al.} later introduced \textit{Deepzzle}~\cite{deepzzle2020tip}, a method for solving $3 \times 3$ puzzles with significant erosions between fragments. The approach first computes a probability matrix representing relative piece placements around a fixed center piece. Based on these predictions, a graph is constructed, and the puzzle is reassembled by solving the shortest path problem.  

Li \textit{et al.} proposed \textit{JigsawGAN}~\cite{jigsawgan2021tip}, a GAN-based learning framework for puzzle solving without relying on ground truth data. The architecture comprises two key components: (1) a classification stream that predicts the correct piece permutation for a shuffled input, and (2) an auxiliary GAN module that generates realistic images. These components are connected through a warp module that corrects classification outputs, enabling the model to emphasize both boundary details between pieces and the global semantic structure of the image. Their experiments focused on solving $4 \times 4$ Type-1 puzzles.  

Talon \textit{et al.} introduced \textit{GANzzle}~\cite{ganzzle2022icip} to reconstruct images from unordered pieces. Their pipeline uses an encoder and a GAN-based module to synthesize a preliminary image. Various target slots are cropped, and embeddings for these targets and the unordered pieces are computed to generate a cost matrix linking patches to target slots. The final piece permutation is determined using a \textit{Hungarian attention} mechanism~\cite{hungarianattention2020}. Despite its innovative design, \textit{GANzzle} was tested only on small-scale Type-1 puzzles and demonstrated relatively low reconstruction accuracy in most reported cases.  

Song \textit{et al.} proposed the \textit{Siamese-discriminant deep reinforcement learning} ($\rm{SD}^2\rm{RL}$) framework~\cite{dqn_jpp2023aaai}, which integrates a Siamese discriminant network with a \textit{deep Q-network} (DQN) trained via reinforcement learning. The system identifies the optimal sequence of fragment swaps for puzzle reassembly, focusing on $5 \times 5$ puzzles with large eroded boundaries.  

Most recently, Liu~\etal~\cite{liu2024diffusion} introduced \textit{JPDVT}, leveraging \textit{vision transformers} (ViTs) to address the challenges of both image and video jigsaw puzzles. Their framework employs conditional generative \textit{diffusion models} for sorting and inpainting, enabling the prediction of positional encodings and the reconstruction of puzzles with missing pieces or video frames. Their results highlight performance on $3 \times 3$ image puzzles and video puzzles with up to 32 frames.  

\subsection{Real-World Visual Reconstruction}
In principle, the generative DL-based CMs and end-to-end (E2E) modules discussed earlier offer a promising alternative to traditional two-phase approaches for solving real-world puzzles, thanks to their adaptability to complex imagery. However, most of these methods fall beyond the scope of this paper due to their limited scalability for larger puzzles, insufficient handling of Type-2 variants, and other constraints. Therefore, our proposed combination of generic DL-based CMs with an enhanced solver, designed for large-scale applications, is particularly relevant for practical puzzle reconstruction in diverse real-world domains.

We now review several real-world problems of interest, revisiting existing methods that have been proposed to address them.

\subsubsection{Portuguese Tile Panels}
The Portuguese tile panel problem~\cite{de2011azulejos} involves the reconstruction of ancient 2D square-tile panels that were removed from various buildings and landmarks in Portugal and its former colonies. Over 100,000 such tiles are currently stored in the Museu Nacional do Azulejo (MNAz) in Lisbon, Portugal, awaiting manual assembly by human experts. Given the extraordinary complexity of this task, it would require decades to complete the assembly of these jigsaw-like puzzles at the current pace~\cite{pais18}.

Fonseca~\cite{fonseca2012montagem} proposed an augmented Lagrange multipliers technique alongside a greedy approach to address Type-1 and Type-2 variants of square-tile puzzles, achieving accuracy rates of 57.8\% and 39.1\%, respectively, on panels containing a few dozen tiles. By comparison, Gallagher’s method~\cite{gallagher2012jigsaw} yielded improved results, with corresponding accuracy rates of 64.5\% and 49.4\%. Andalo {\etal}~\cite{andalo2016impa} achieved perfect reconstruction for four mixed tile panels using their PSQP method~\cite{7442162}, provided tile orientations were known. However, their approach did not address Type-2 puzzles and was limited to relatively small, high-resolution panels.

More recently, Rika {\etal}~\cite{rika2019gecco} introduced a DL-based approach combined with a genetic algorithm (GA) solver, producing promising preliminary results. The methodology presented in this paper builds upon and significantly extends their work.

\subsubsection{Eroded Boundaries}
\begin{figure*}[t!]
    \centering
    \subfloat[]{\includegraphics[width=0.3\linewidth]{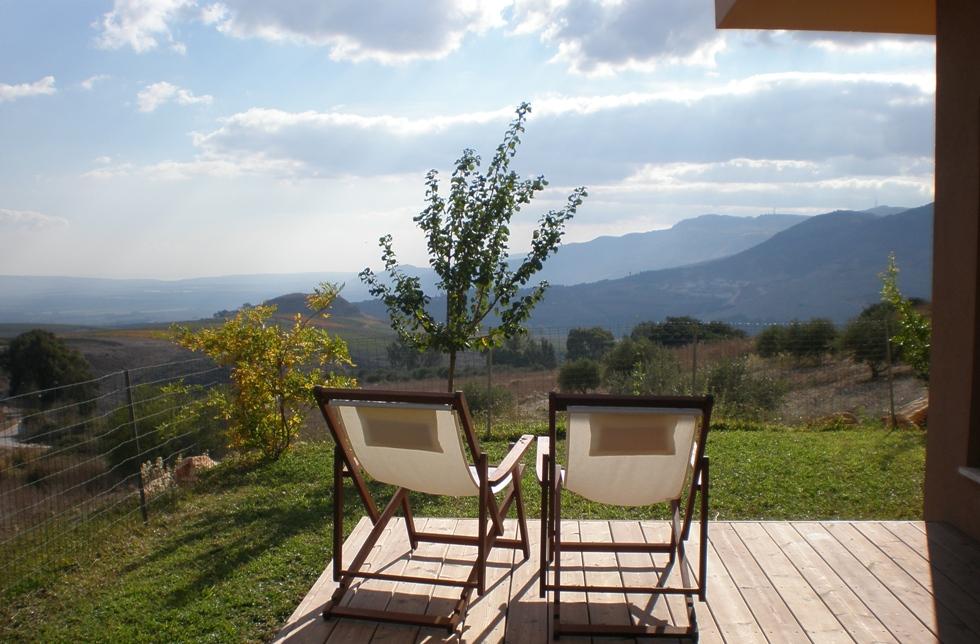}}
    \hfill
    \subfloat[]{\includegraphics[width=0.3\linewidth]{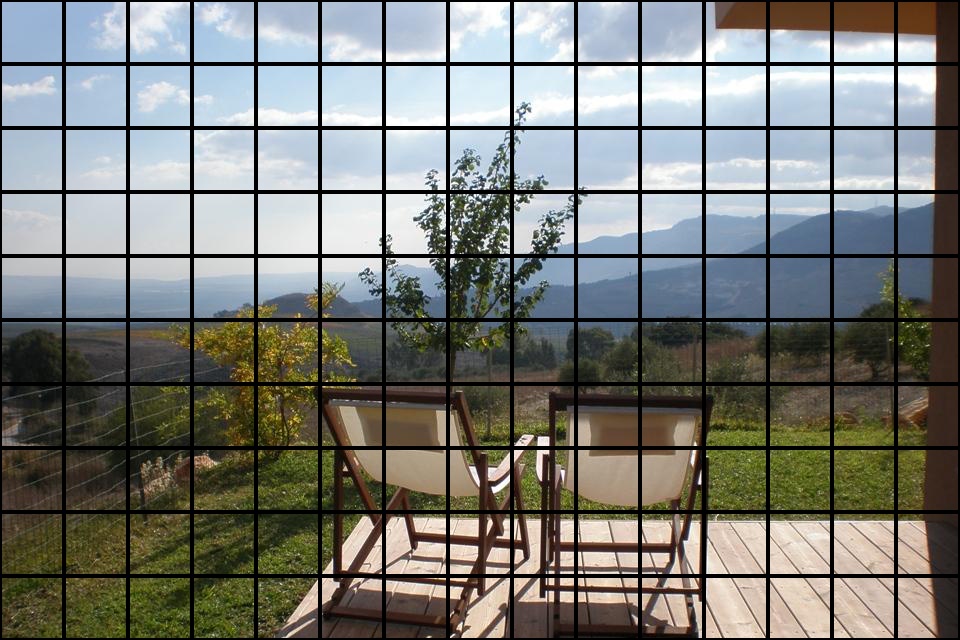}}
    \hfill
    \subfloat[]{\includegraphics[width=0.3\linewidth]{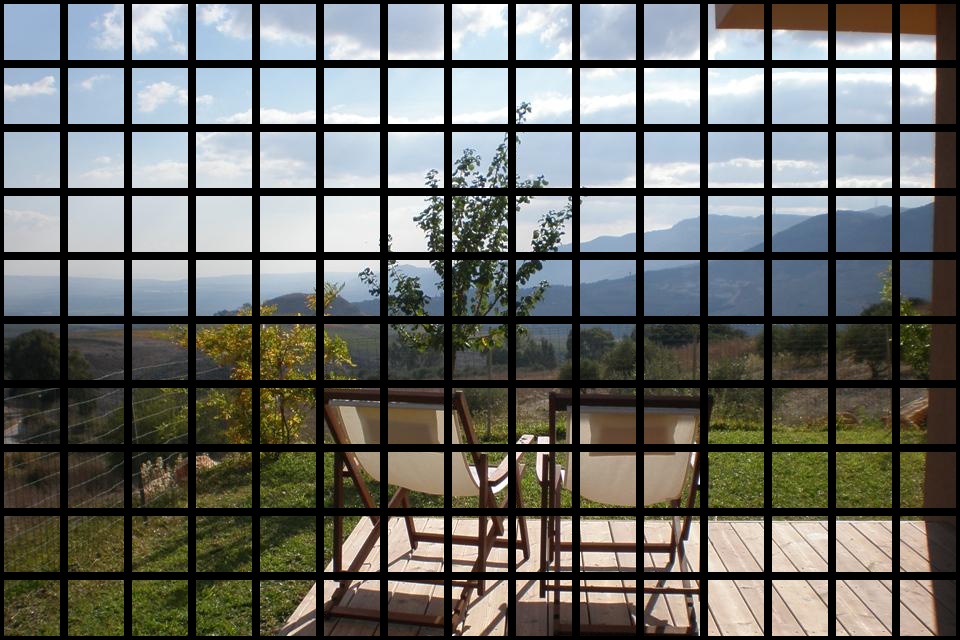}}
    \caption{Artificially eroded boundaries of puzzle with 150 $64 \times 64$ square tiles: \textbf{(a)} Original image ($t=0$), \textbf{(b)} image with 2-pixel erosion layers ($t=2$), and \textbf{(c)} image with 4-pixel erosion layers ($t=4$).}
    \label{fig:eroded_boundaries}
\end{figure*}

To simulate eroded boundaries, $t$-pixel layers around the edges of each tile are replaced with black (\ie, zero) pixels. Figure~\ref{fig:eroded_boundaries} illustrates an example puzzle with varying erosion widths of $t$.

Mondal {\etal}~\cite{mondal2013crv} were among the first to address eroded boundaries using an enhanced \textit{selectively weighted MGC} (wMGC) compatibility measure, which combines SSD and MGC dissimilarities. While wMGC achieved modest improvements in reconstruction, its accuracy dropped sharply as $t$ increased: from 90\% at $t=0$ to 54\% at $t=1$, and 38\% at $t=2$.

Bridger {\etal}~\cite{bridger2020cvpr} proposed a GAN-based solution for eroded boundaries, integrated with the greedy solver from~\cite{paikin2015solving}. To ensure GAN convergence, the tile size was increased from $28 \times 28$ to $64 \times 64$, with erosion widths of $t=\{2,4\}$. They achieved neighbor accuracy rates of 79.3\% and 53.2\% for $t=2$ and $t=4$, respectively.

Khoroshiltseva {\etal}'s JiGAN~\cite{JiGAN2022iciap} also utilized high-resolution $64 \times 64$ tiles. While more computationally efficient than Bridger {\etal}, JiGAN demonstrated significantly lower performance.

\subsubsection{Shredded Documents}
Skeoch~\cite{skeoch2006} explored various dissimilarity measures and color models for compatibility evaluation. However, these methods are less relevant for text documents, which are predominantly black and white.

Ranca~\cite{ranca2013} developed a probabilistic model to estimate adjacency likelihoods for pairs of shreds, using a variant of the minimum spanning tree (MST) algorithm to reconstruct shredded documents. This method achieved nearly perfect reconstruction for some datasets and an average accuracy of roughly 50\% for strip- and cross-cut shredded documents.

More recently, Paixao {\etal}~\cite{paixao2018deep, paixão2020faster} introduced DL-based compatibility measures for strip-cut shredded documents. By employing an ATSP optimizer, they achieved state-of-the-art results for this problem.

\section{Deep Learning-Based Compatibility Measure (DLCM)}
\label{sec:cnn_based_cm}
We propose a {\em deep learning compatibility measure} (DLCM) to achieve a precise and resilient CM tailored for real-world, puzzle-related tasks involving non-overlapping square pieces. The DL component is implemented using a compact {\em convolutional neural network} (CNN) that processes an entire pair of pieces and outputs a scalar $s \in \mathbb{R}$. The value of $s$ correlates with the likelihood of adjacency between the pair in the original image. Ideally, the DLCM satisfies:
\begin{equation}\label{eq:triplet_citiria}
    \forall e_n \neq e_p: f(e_a, e_p) > f(e_a, e_n),
\end{equation}
where $f$ represents the DLCM, $e_a$ and $e_p$ are, respectively, the anchor piece boundary and its true neighbor, and $e_n$ denotes any non-neighboring boundary. The architecture, training, and post-processing of our DLCM are detailed below.

\subsection{Triplet Selection}
\label{sec:dlcm-2}

Based on Eq.~\ref{eq:triplet_citiria}, the DLCM training set is constructed from triplets of the form $(e_a, e_p, e_n)$. Suitable objective functions include the {\em triplet-loss}:
\begin{equation}\label{eq:triplet_loss}
    \begin{split}
       {\mathcal{L}_{\text {Triplet Loss}}} = {\rm {max}}(0, \gamma - f(e_a,e_p) + f(e_a,e_n)),
    \end{split}
\end{equation}
and the {\em binary cross entropy} (BCE) loss:
\begin{equation}\label{eq:bce_loss}
    {\mathcal{L}_{\text {BCE}}} = -\left[ \log(\sigma(f(e_a, e_p)))+\log(1 - \sigma(f(e_a, e_n))) \right].
\end{equation}
Here, $f$ denotes the DLCM score, $\sigma$ is the sigmoid function, and $\gamma$ is a hyperparameter fixed at $1$. Experiments revealed that BCE outperforms the triplet-loss in both convergence and generalization, as discussed in Section~\ref{sec:experimental_results}.

While DLCM training employs {\em logistic regression} to interpret output as probabilities of positive or negative pairs, the sigmoid function is omitted during inference to leverage the wider raw compatibility score range for robust pair assessment.

Given the extensive possible triplets, training triplets are sampled online. Specifically, piece boundaries from a randomly selected image serve as anchors, generating positive pairs with adjacent edges (typically four per piece, fewer at boundaries or corners). Negative pairs are formed by pairing anchors with non-adjacent edges from the dataset.

\begin{figure}[!t]
\centering
    \subfloat[]{\includegraphics[width=0.3\columnwidth]{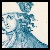}}
    \hfill
    \subfloat[]{\includegraphics[width=0.3\columnwidth]{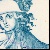}}
    \hfill
    \subfloat[]{\includegraphics[width=0.3\columnwidth]{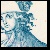}}
\caption{Piece augmentation through degradation and shift: \textbf{(a)} Tile degraded by removing a 2-pixel frame, \textbf{(b)} tile shifted one pixel to the left and one pixel upward, and \textbf{(c)} combined augmentation of (a) and (b).}
\label{fig:augmented_tiles}
\end{figure}

To address data insufficiency or imperfections, we augment triplets by applying {\em degradation} and/or {\em shifts}. Degradation substitutes random boundary pixels with zeros, replacing no pixels, single-pixel boundaries, or double-pixel frames uniformly. This improves the network's learning focus beyond edge textures. Shifts displace pieces randomly by zero to two pixels horizontally or vertically, filling gaps with zeros. Figure~\ref{fig:augmented_tiles} illustrates these augmentations. Notably, these augmentations were beneficial for the Portuguese tile panels problem but were excluded from training in other domains.

\subsection{Deep CNNs}\label{sec:cnn_model}

We target square-piece puzzles, {\ie}, tiles of size $P \times P$ pixels, with $P$ determined by the dataset. For the Portuguese tile panels, high-resolution tiles from the MNAz were downscaled to $50 \times 50$ pixels. The CNN model processes essentially concatenated piece pairs of size $P \times 2P$ to evaluate the CM. This concatenated input format effectively ``coerces'' the model to simultaneously consider both puzzle pieces as a unified entity, enabling it to learn and assess their degree of compatibility in a more coherent and robust manner.

Anchors are consistently placed on the left, with the adjacent piece edges rotated as needed to align for compatibility comparison. For example, left edges of anchors are matched to the right edges of other pieces by rotating both by 180\degree.

\begin{figure*}[!t]
    \centering
    \includegraphics[width=0.8\linewidth]{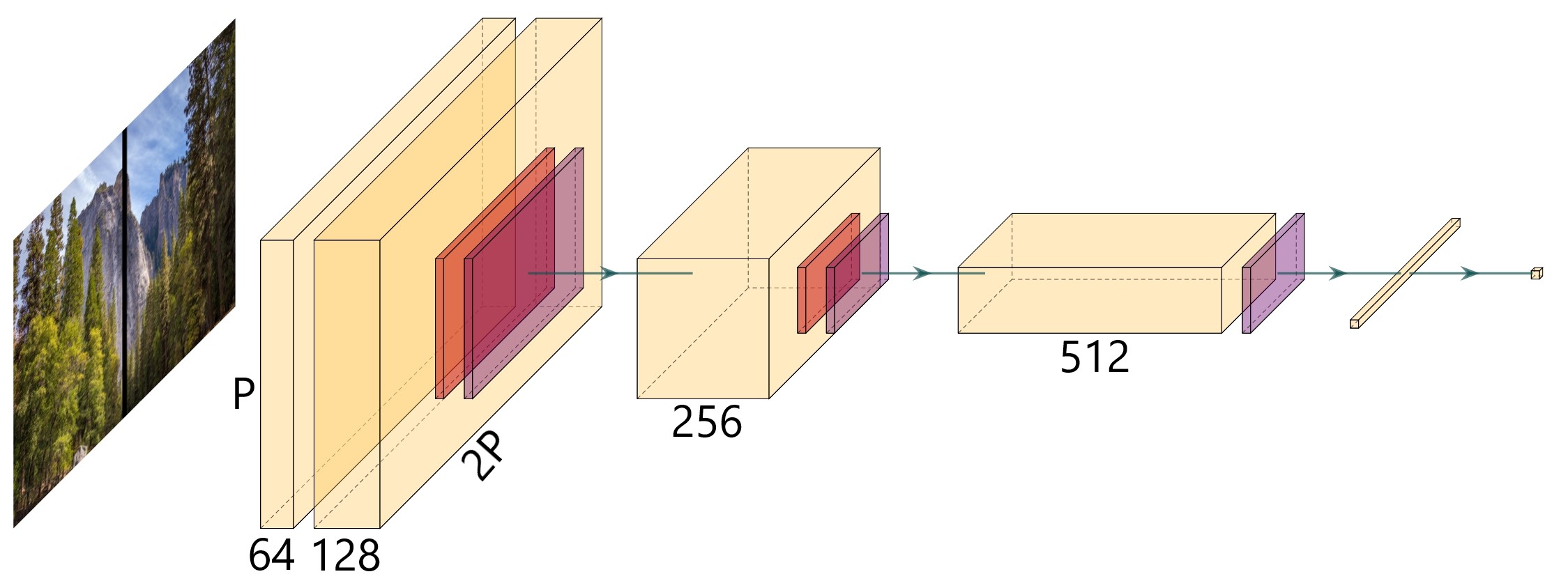}
    \caption{Sub-model architecture with input size $P \times 2P \times C$, where the piece size is $P \times P$ pixels and $C$ is the number of channels; the architecture includes four convolutional layers with $3 \times 3$ kernels and ReLU activation function; max pooling is applied after the second and third layers, and dropout with a probability of 0.25 is applied after all layers except the first; the final convolutional layer is flattened and passed through a fully-connected layer to compute the compatibility score without an activation function; no biases are used in any layer.}
    \label{fig:atom_architecture}
\end{figure*}

The DLCM architecture for the Portuguese tile problem comprises four sub-models: Red-Net, Green-Net, Blue-Net, and RGB-Net. Each follows the architecture in Figure~\ref{fig:atom_architecture}.

\begin{figure}[!t]
    \centering
    \includegraphics[width=0.9\columnwidth]{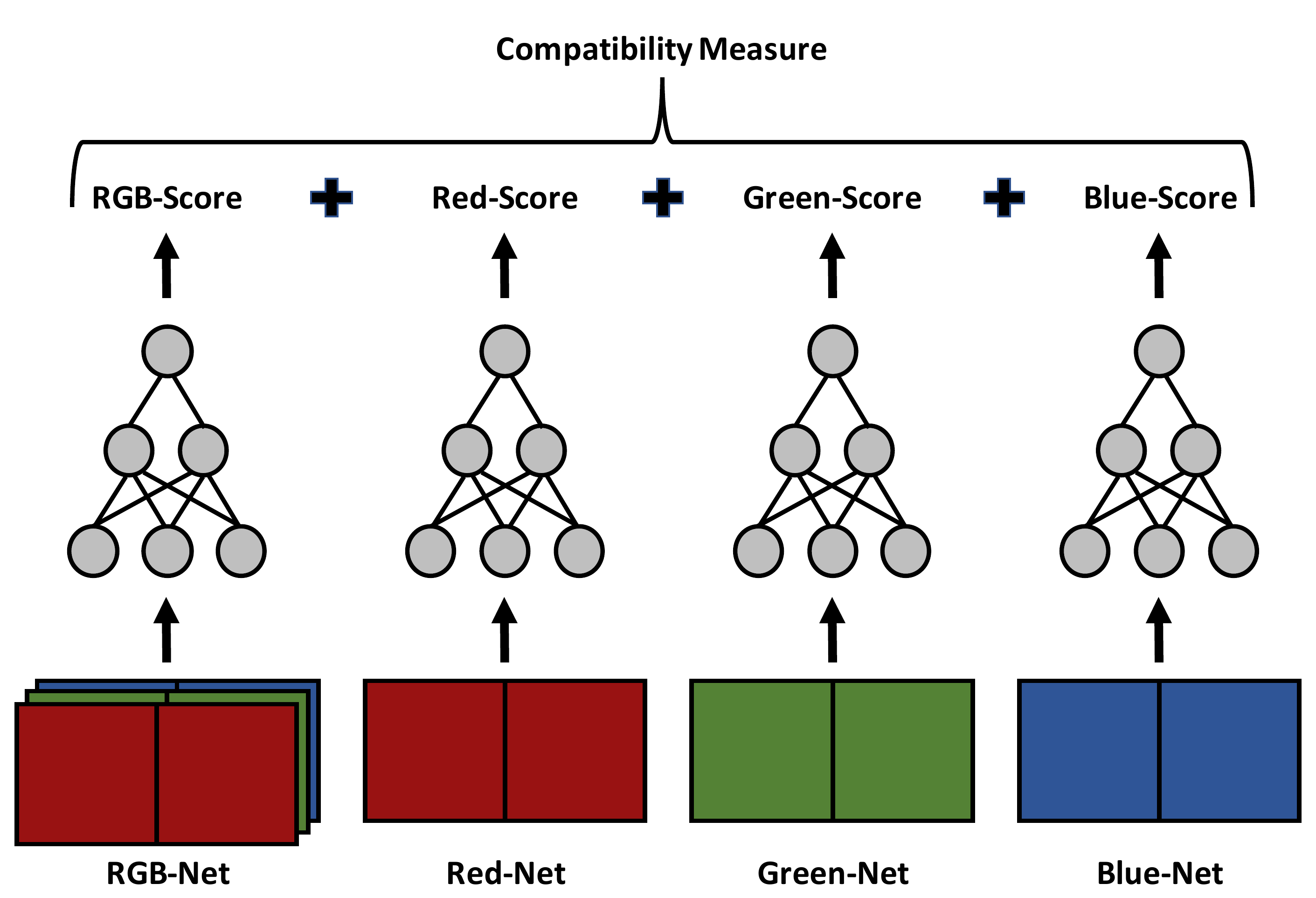}
    \caption{Our DLCM architecture, consisting of four sub-models: RGB-Net, Red-Net, Green-Net, and Blue-Net, with the same architecture as Figure~\ref{fig:atom_architecture}; the DLCM output is the sum of the outputs of all four sub-models.}
    \label{fig:DLCM_architecture}
\end{figure}

The complete DLCM ensemble architecture is shown in Figure~\ref{fig:DLCM_architecture}. Given the monochromatic nature of Portuguese tiles, training separate models for Red, Green, and Blue channels enhances performance, while RGB-Net captures inter-channel dependencies. Each sub-model is trained independently on identical batches, with isolated loss computations.

\subsection{Post Processing}\label{sec:post_processing}

Global min-max normalization is applied per edge to scale compatibility scores between $0$ and $1$:
\begin{equation}
    C'(e_i,e_j) = \dfrac{C(e_i,e_j) - min(C(e_i,*))}{max(C(e_i,*)) - min(C(e_i,*))}.
\end{equation}
As symmetry discrepancies ($C(e_i,e_j) \neq C(e_j,e_i)$) are undesirable, symmetry is ensured via averaging:
\begin{equation}
    C''(e_i,e_j) = C''(e_j,e_i) = \dfrac{C'(e_i,e_j) + C'(e_j,e_i)}{2}.
\end{equation}

Section~\ref{sec:experimental_results} discusses the resulting CM's performance, demonstrating its robustness and high accuracy across various real-world domains.

\section{GA Solver}
\label{sec:ga_solver}

Previous greedy solvers often rely on heuristic methods to minimize incorrect placements during the initial stages of reconstruction. Although these solvers generally yield true edge adjacencies with high probability, they remain susceptible to nonrecoverable assignment errors. 

To address this limitation and reduce the reliance on ad-hoc criteria, we adopt the GA-based solver proposed by Sholomon {\etal}~\cite{Sholomon_2013_CVPR}. The stochastic nature of genetic algorithms allows for the correction of erroneous adjacencies during global optimization, improving reconstruction outcomes.

The core components of a GA~\cite{holland1975adaptation} include a \textit{population} of candidate solutions (\ie, \textit{chromosomes}), a \textit{fitness function} to assess the quality of these solutions, and genetic operators such as \textit{crossover} and \textit{mutation} to generate new offspring (\ie, new solutions). The population evolves through successive \textit{generations}, with selection mechanisms ensuring that higher-quality solutions have a greater likelihood of propagating their characteristics. Over time, this iterative process steers the algorithm toward better solutions.

Our GA-based solver includes two essential elements: (1) a {\em fitness function} to evaluate the quality of proposed puzzle reconstructions and (2) a {\em crossover operator} that combines two tile configurations (\ie, parent chromosomes) to produce a new configuration (\ie, offspring), ideally improving upon the parents.

The fitness function is defined as the sum of pairwise compatibility scores across all reconstructed puzzle boundaries. Our crossover operator builds upon the hierarchical design introduced in~\cite{Sholomon_2013_CVPR}, incorporating enhancements such as the addition of a \textit{best-buddies} phase inspired by~\cite{Sholomon_2013_CVPR}. This phase enforces the placement of two tiles together only if they are mutually the most compatible. We have integrated this phase as Phase 3 of our hierarchical scheme, leveraging insights from~\cite{rika2019gecco} and practical applications.

In our crossover operator, tiles are added to the kernel iteratively based on the following hierarchical scheme, which terminates when all puzzle pieces are included in the kernel:

\begin{itemize}
    \item \textbf{Phase 1.1:} Add a neighboring piece (relative to the free boundary of the kernel) from a parent chromosome if its compatibility score exceeds $\alpha=\max(\alpha_{0}, C_{\rm{mean}})$, where $C_{\rm{mean}}$ is the average compatibility score for the chromosome, and $\alpha_{0} \in (0,1)$ is an initial threshold. In our experiments, $\alpha_{0}=0.8$ yielded the best results. The score of a piece is defined as the average compatibility measure (CM) with all its neighbors. This phase prioritizes the parent chromosome with the higher fitness to ensure accurate reconstruction.
    
    \item \textbf{Phase 1.2:} Similar to Phase 1.1, but selects the parent chromosome with the lower fitness.

    \item \textbf{Phase 2:} Place a piece if both parent chromosomes agree on its adjacency to the kernel.

    \item \textbf{Phase 3:} Add a piece if it forms a best-buddy pair (mutually most compatible) with a piece on the kernel's free boundary in one of the parent chromosomes.

    \item \textbf{Phase 4.1:} Place the most compatible piece available with respect to a free boundary of the kernel.

    \item \textbf{Phase 4.2:} Place the second-most compatible piece available with respect to the free boundary.

    \item \textbf{Phase 5:} Randomly select one of the remaining pieces and place it at a free boundary of the kernel.
\end{itemize}

\begin{figure*}
    \centering
    \includegraphics[width=0.9\linewidth]{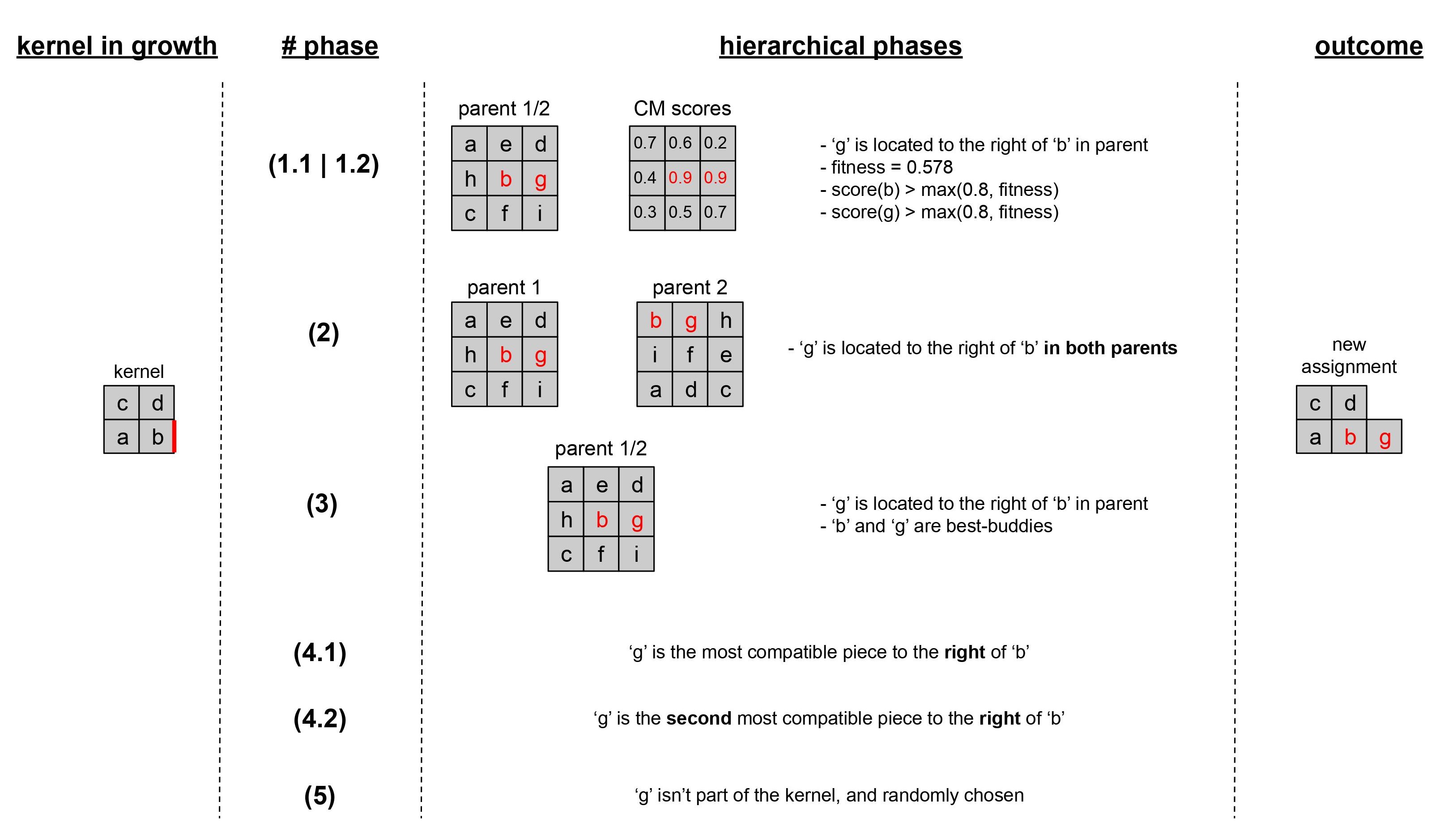}

\caption{Illustration of tile placement through chromosome pairing in different phases of the proposed genetic algorithm (GA); starting with a single piece, the kernel expands by sequentially placing additional tiles adjacent to its free (piece) edges. In the depicted toy scenario, the kernel consists of four (out of nine possible pieces) while specifically examining the right boundary of tile 'b'. \textbf{Phase 1.1/1.2:} For the left parent chromosome, the average CM scores of pieces 'b' and 'g' exceed the chromosome's overall fitness; \textbf{Phase 2:} Both parent chromosomes share the adjacency of piece 'g' to the right of piece 'b'; \textbf{Phase 3:} Pieces 'b' and 'g' are best buddies in the left parent; \textbf{Phases 4.1, 4.2,} and \textbf{Phase 5:} see details in the figure itself. The new kernel (and resulting offspring chromosome), due to any of the above phases, includes piece 'g' to the right of piece 'b'.}

    \label{fig:GA_phases}
\end{figure*}

Figure~\ref{fig:GA_phases} illustrates these hierarchical phases of our crossover operator. 

The sequence of GA phases was determined empirically through extensive experimentation. Subsection~\ref{subsec:GA_reconstruction} provides additional details on the choice of hyperparameters for our GA and its comparative performance in reconstruction tasks.

\section{Empirical Studies}
\label{sec:experimental_results}
This section provides a comparative performance evaluation between several established methods and our generic hybrid scheme. The results presented here primarily focus on the Portuguese tile data, as this problem domain incorporates many of the challenging aspects typical of 2D real-world reconstruction, {\eg}, degraded content information due to eroded boundaries and faded coloring, missing pieces, multiple puzzles, etc. (A broader comparative evaluation for additional related problem domains is provided in Section~\ref{sec:other_problem_domains}.) Unless explicitly stated otherwise, the DLCM evaluation and reconstruction results in this section pertain to the 24-panel test set provided by the MNAz of Lisbon, Portugal.

All networks were trained using \textit{stochastic gradient descent} (SGD) with standard backpropagation~\cite{rumelhart1986learning} and the Adam optimizer~\cite{DBLP:journals/corr/KingmaB14}, with a learning rate of 0.0001 and a batch size of 64. Training was conducted on a modern PC with a 3.5GHz CPU, 32GB RAM, and a single GPU with 11GB of memory. A detailed discussion of experimental results is provided in Section~\ref{sec:experimental_results}.

\subsection{Portuguese Tile Datasets}
\label{sec:datasets}
The 24 high-resolution test images from the MNAz were excluded from the CNN training process. (As described in Subsection~\ref{sec:cnn_model}, all tiles were downscaled to $50 \times 50$ pixels.) The largest image contains 460 tiles, while the smallest comprises 90 tiles, with an average of 201 tiles per test image.

We also acquired nine smaller images from the MNAz: five with 25 pieces each, and four containing 40, 48, 60, and 72 pieces, respectively. Due to their relatively small size, these images may not perfectly represent the complexities of the reconstruction problem. Nevertheless, given their source, they were considered sufficiently representative in terms of content and were used without reservation as a held-out validation set during CNN training.

Additionally, we collected 217 images of Portuguese tile panels from the Internet, some of which were taken by casual tourists. Each puzzle was manually inspected to determine the number of pieces per row and column. Based on the image dimensions, all were resized to $50 \times 50$ pixels. Nine of these images were manually cropped along tile lines and added to the validation set, while the remaining 208 images were used for CNN training. Following automated piece-cropping, the dataset contained 22,237 pieces. Although automated cropping may not always align perfectly with actual piece boundaries, such occurrences are rare and may contribute positively by reducing the risk of overfitting.

In summary, the CM model was trained on a dataset of 208 images, with a validation set comprising 18 images (nine from the MNAz and nine sourced online). For evaluating the CM model and overall reconstruction accuracy, the 24 high-resolution test images provided by the MNAz were utilized.

\subsection{Compatibility Measure Evaluation}  

\begin{figure}[!t]
    \centering
    
    \tabular{c}
    \includegraphics[width=0.75\columnwidth]{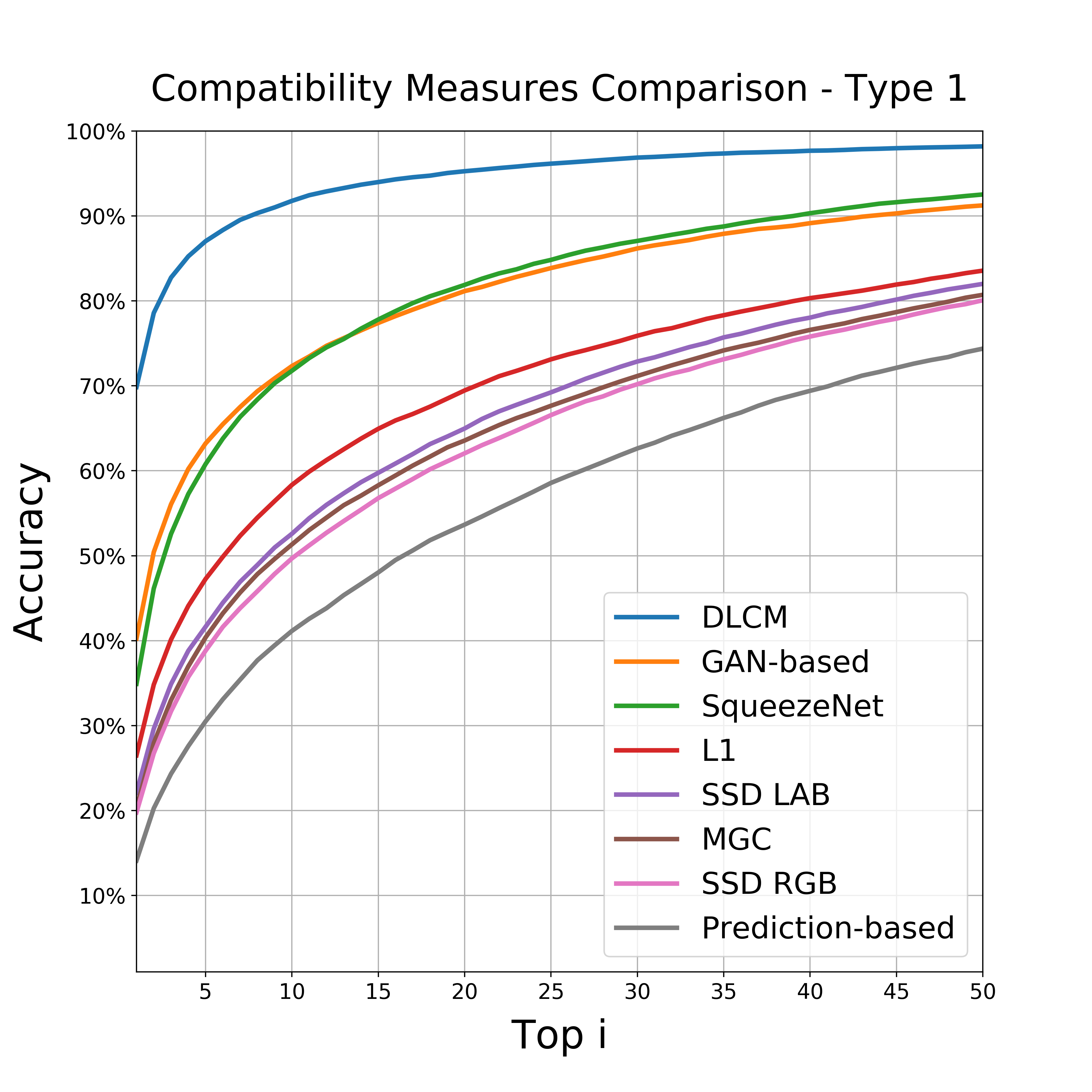} \\
    \includegraphics[width=0.75\columnwidth]{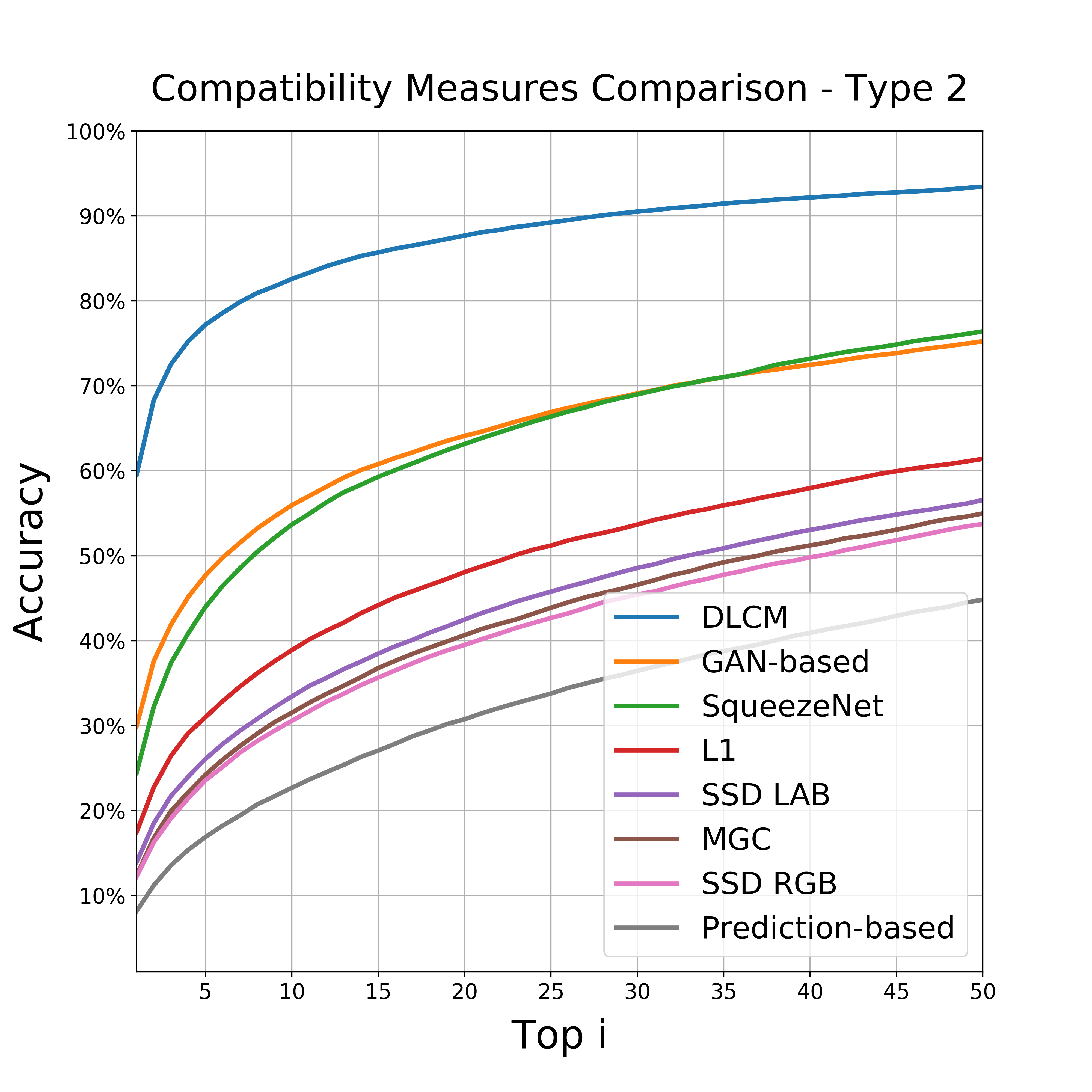}
    \endtabular
    
    \caption{Top-$i$ accuracy plots comparing traditional and DL-based CMs with the proposed DLCM for Type-1 (top) and Type-2 (bottom) Portuguese tile panels; DLCM achieves superior performance over all tested methods, including SqueezeNet- and GAN-based CMs.}
    \label{fig:top_i_curve}
\end{figure}

\begin{enumerate}  
\item  
\textit{Top-$i$ Measure}:  
\newline  
To assess the accuracy of a CM, we utilized the Top-$i$ measure, which is defined as the likelihood that a positive pair ranks (at least) as the $i$-th most compatible among all potential (piece boundary) candidates for a given tile boundary. By applying this definition to all piece boundaries of a puzzle, we compute the Top-$i$ measure for that puzzle.  
To calculate the Top-$i$ measure for a set of puzzles, we take a (weighted) average of the individual Top-$i$ measures for each puzzle in the set.  

The above definition ensures that ${\rm {Top}}$-$i \ge {\rm {Top}}$-$(i-1)$. Furthermore, the ideal Top-$1$ accuracy of a CM should approach $100\%$, meaning that the most compatible tile (for a given piece boundary) forms a positive pair with the piece sharing that boundary. Therefore, Top-$1$ is the most important metric for evaluating a CM's performance, as shown in Table~\ref{tab:cm_compare}.  

We trained the DLCM as previously described and evaluated it on the test set outlined in Subsection~\ref{sec:datasets}.  
Our DLCM achieves Top-$1$ accuracy rates of 69.9\% and 59.5\% on Type-1 and Type-2 puzzles, respectively. Without the post-processing described in Subsection~\ref{sec:post_processing}, the DLCM achieves Top-$1$ accuracy rates of only 64.5\% and 54.3\%, respectively, for these two puzzle variants. This clearly highlights the significant contribution of our proposed post-processing approach.  

To evaluate the individual contribution of each of the four DLCM sub-networks, we also computed their Top-$1$ accuracy. The results are provided in Table~\ref{tab:cm_compare} for reference. These findings indicate that no single sub-network outperforms the DLCM ensemble, strongly suggesting that each sub-network captures unique features. The combined operation of these sub-networks leads to enhanced overall performance. For visualizations of the DLCM's performance, see Appendix~\ref{sec:compatibility map}.  

\item  
\textit{Top-$i$ Accuracy of Various CMs}:  
\newline  
We performed a comprehensive comparison among several CMs, including traditional measures applied only to neighboring piece columns, as well as more advanced DL-based measures leveraging full-content information. Specifically, we compared our DLCM to the chromatic SSD~\cite{conf/cvpr/PomeranzSB11}, which achieves Top-$1$ accuracy rates of 19.8\% and 12.2\% on Type-1 and Type-2 puzzles, respectively, and to MGC~\cite{gallagher2012jigsaw}, which achieves Top-$1$ accuracy rates of 20.3\% and 12.3\% for these variants. While more advanced DL-based models, such as the SqueezeNet-based V1.1 model~\cite{paixao2018deep} and the GAN-based model from~\cite{bridger2020cvpr}, outperform traditional methods, their Top-$1$ accuracy remains significantly lower than that of our DLCM, even after retraining them on the same Portuguese panel dataset with favorable resolutions. (See Appendix~\ref{sec:other_DL_CMs} for details on retraining these models on the Portuguese tile dataset.)  

A detailed summary of the comparative study is presented in Table~\ref{tab:cm_compare} and Figure~\ref{fig:top_i_curve}.  

\end{enumerate}  

\begin{table}[!t]
    \centering
    \caption{Top-1 accuracy of traditional and DL-based CMs compared to the proposed DLCM sub-networks and DLCM ensemble, evaluated on Type-1 and Type-2 Portuguese tile panels.}
    
    \begin{tabular}{lcc}
        \toprule
        \multirow{2}{*}{\textbf{Compatibility Measure}} & \multirow{2}{*}{\textbf{Type-1}} & \multirow{2}{*}{\textbf{Type-2}} \\
         &  &  \\
        \midrule
        Prediction-based \cite{conf/cvpr/PomeranzSB11} & 14.1\% & 8.1\% \\
        SSD (RGB) \cite{conf/cvpr/PomeranzSB11} & 19.8\% & 12.2\% \\
        MGC \cite{gallagher2012jigsaw} & 20.3\% & 12.3\% \\
        SSD (LAB) \cite{conf/cvpr/PomeranzSB11} & 22.1\% & 13.9\% \\
        $L_1$ \cite{paikin2015solving} & 26.5\% & 17.4\% \\
        SqueezeNet-based \cite{paixao2018deep} & 34.9\% & 24.4\% \\
        GAN-based \cite{bridger2020cvpr} & 40.3\% & 29.9\% \\
        \midrule
        \multicolumn{3}{c}{\textbf{Proposed}} \\
        \midrule
        Red-Net & 64.1\% & 53.2\% \\
        Green-Net & 64.6\% & 53.7\% \\
        Blue-Net & 60.4\% & 48.9\% \\
        RGB-Net & 62.5\% & 51.6\% \\
        \textbf{DLCM} & \textbf{69.9\%} & \textbf{59.5\%} \\
        \bottomrule
    \end{tabular}
    \label{tab:cm_compare}
\end{table}

The above comparison clearly indicates the superiority of our DLCM model to various established CMs, including other DL-based measures.
Finally, the DLCM results reported in Table~\ref{tab:cm_compare} exceed those in~\cite{rika2019gecco} by 1.5\% and 3.4\% for Type-1 and Type-2, respectively.

\subsection{GA-Based Reconstruction}
\label{subsec:GA_reconstruction}
The order of GA phases, as well as its  various hyper-parameters, were determined by extensive trial and error experimentation. To preserve the best attributes along the evolutionary process, the best chromosome at each level is passed on to the next generation ({\ie},  \textit{elitism} is set to 1). Also, chromosome selection at each stage is carried out due to the \textit{roulette wheel selection} procedure to reflect a chromosome weight according to their proportionate fitness. In an attempt to escape local maxima, the GA solver introduces additional randomness to the reconstruction process (via mutation), by skipping some of the crossover phases (of Section~\ref{sec:ga_solver}) with small probability. Specifically, the GA skips Phases 1.1 and 1.2, with 10\% probability, and Phases 2 and 3, with 20\% probability. The other phases remain intact. In addition, the population in each generation was set to 100 chromosomes, and the GA terminates upon failing to obtain an improved chromosome over 50 consecutive generations.

We incorporated our newly trained DLCM into a modified GA framework, in an attempt to reconstruct each of the test set images. We report below the reconstruction accuracy, according to the {\em neighbor comparison} definition used in previous works, {\ie}, the fraction of correctly assigned adjacent edges with respect to ground truth.

We attempted reconstruction for four problem variants, with unknown piece location common to all of them. The variants differ as to a priori knowledge of piece orientation and puzzle dimensions. The hardest variant is the one for which both piece orientation and puzzle dimensions are unknown.

We report the best result, after running our enhanced GA module 50 times on each image. For comparison, we reconstructed the images according to~\cite{conf/cvpr/PomeranzSB11} and~\cite{gallagher2012jigsaw}, since these methods are very common and since their code is publicly available. Also, we implemented the reconstruction used by Paikin {\etal}~\cite{paikin2015solving} and Bridger {\etal}~\cite{bridger2020cvpr}, claimed as SOTA in their associated problem domains, and evaluated them on the Portuguese panel problem. A comparative reconstruction summary 
is given in Table~\ref{tab:reconstruction_results}.
Interestingly, the methods compared against return usually worse reconstruction accuracy for known dimensions. This could be attributed to the fact that relatively inferior CMs tend to exceed more often the original puzzle dimensions, in an attempt to improve the overall accuracy, thereby producing ill-shaped puzzle configurations.

\subsection{GA-Based Reconstruction}  
\label{subsec:GA_reconstruction}  
The order of GA phases and its various hyper-parameters were determined through extensive trial-and-error experimentation. To retain the best attributes throughout the evolutionary process, the best chromosome at each level is passed to the next generation ({\ie}, \textit{elitism} is set to 1). Additionally, chromosome selection at each stage is conducted using the \textit{roulette wheel selection} procedure, which reflects chromosome weights based on their proportionate fitness.  

To avoid local maxima, the GA solver introduces randomness into the reconstruction process (via mutation) by skipping some of the crossover phases (as outlined in Section~\ref{sec:ga_solver}) with a small probability. Specifically, Phases 1.1 and 1.2 are skipped with a 10\% probability, and Phases 2 and 3 are skipped with a 20\% probability. All other phases remain unchanged. Furthermore, the population size in each generation is set to 100 chromosomes, and the GA terminates if no improved chromosome is found over 50 consecutive generations.  

We integrated our newly trained DLCM into a modified GA framework to reconstruct each of the test set images. Below, we report the reconstruction accuracy based on the {\em neighbor comparison} metric used in previous studies, {\ie}, the fraction of correctly assigned adjacent edges relative to the ground truth.  

We performed reconstruction for four problem variants, all sharing the characteristic of unknown piece locations. These variants differ in terms of prior knowledge of piece orientation and puzzle dimensions. The most challenging variant involves unknown piece orientation and unknown puzzle dimensions.  

We report the best results after running our enhanced GA module 50 times on each image. For comparison, we reconstructed the images using the methods proposed by~\cite{conf/cvpr/PomeranzSB11} and~\cite{gallagher2012jigsaw}, as these are widely used and their code is publicly available. We also implemented the reconstruction methods by Paikin {\etal}~\cite{paikin2015solving} and Bridger {\etal}~\cite{bridger2020cvpr}, which are claimed as SOTA for their respective problem domains, and evaluated them on the Portuguese panel problem. A comparative reconstruction summary is provided in Table~\ref{tab:reconstruction_results}.  

Interestingly, the methods compared tend to yield worse reconstruction accuracy for puzzles with known dimensions. This may be due to relatively inferior CMs more frequently exceeding the original puzzle dimensions in an effort to maximize accuracy, resulting in ill-shaped puzzle configurations.

\begin{table}[!t]
\centering
\caption{Neighbor accuracy of reconstruction for previous methods compared to the proposed hybrid scheme, applied to Type-1 and Type-2 Portuguese tile panels with known and unknown dimensions.}
\begin{tabular}{|c||c|c|}
    \hline
    \multirow{3}{*}{\textbf{Method}} & \multicolumn{2}{c|}{\textbf{Type-1}} \\
    \cline{2-3}
    & Known & Unknown \\
    & Dims.  & Dims. \\
    \hline
    \hline
    Pomeranz {\etal}~\cite{conf/cvpr/PomeranzSB11} & 10.4\% & - \\
    \hline
    Gallagher~\cite{gallagher2012jigsaw} & 13.7\% & 17.9\% \\
    \hline
    Paikin {\etal}~\cite{paikin2015solving} & 17.2\% & 19.4\% \\
    \hline
    Bridger {\etal}~\cite{bridger2020cvpr} & 28\% & 32.3\% \\
    \hline
    Proposed & \textbf{95.2\%} & \textbf{95.4\%} \\
    \hline
    \hline
    
    \multirow{3}{*}{\textbf{Method}} & \multicolumn{2}{c|}{\textbf{Type-2}} \\
    \cline{2-3}
    & Known & Unknown \\
    & Dims.  & Dims. \\
    \hline
    \hline
    Pomeranz {\etal}~\cite{conf/cvpr/PomeranzSB11} & - & - \\
    \hline
    Gallagher~\cite{gallagher2012jigsaw} & 4.6\% & 4.5\% \\
    \hline
    Paikin {\etal}~\cite{paikin2015solving} & 11.1\% & 12.6\% \\
    \hline
    Bridger {\etal}~\cite{bridger2020cvpr} & 18.5\% & 21.6\% \\
    \hline
    Proposed & \textbf{89.4\%} & \textbf{86.6\%}  \\
    \hline
    \end{tabular}
\label{tab:reconstruction_results}
\end{table}

The bottom-line results, achieved through various modifications along with extensive training and testing, are comparable to or superior to the preliminary Type-1 and Type-2 results reported in~\cite{rika2019gecco}. Specifically, the Type-2 results surpass the previous accuracies by 2.6\% and 4.4\% for puzzles with known and unknown dimensions, respectively~\cite{rika2019gecco}. In summary, our generic method achieved SOTA accuracy on the Portuguese panel problem ({\eg}, 95.2\% and 89.4\% for Type-1 and Type-2 puzzles with known dimensions, respectively).

To further evaluate the contributions of our proposed improvements to the GA-based solver, we conducted an ablation study of its various phases. Specifically, we removed each individual component (other than the essential Phases 4 and 5) and recorded the resulting reconstruction accuracy using the same pairwise CMs. Due to the stochastic nature of the GA, each test was repeated 50 times, and we recorded both the best and average neighbor accuracy on the 24 Portuguese test panels.  

For a comprehensive and fair comparison, we also implemented the original GA-based solver by Sholomon \etal~\cite{Sholomon_2013_CVPR} and executed it with identical pairwise CMs. As shown in Table~\ref{tab:ga_ablation}, our innovative Phase 1 (consisting of Phases 1.1 and 1.2) makes the most significant contribution to our enhanced GA version. The removal of Phases 2 and 3 leads to a relatively smaller degradation in performance. Mutations are also crucial, particularly for the Type-2 problem variant. Additionally, note the substantial performance improvement relative to~\cite{Sholomon_2013_CVPR}.

\begin{table}
\centering
\caption{Ablation Analysis of GA Phases}
\begin{tabular}{l c c c c c}
 
& \multicolumn{2}{c}{\textbf{Type-1}} && \multicolumn{2}{c}{\textbf{Type-2}} \\ \cline{2-3} \cline{5-6}

\textbf{Method} &  &  &  &  &  \\

& \textbf{Avg.} & \textbf{Best} && \textbf{Avg.} & \textbf{Best} \\

\midrule
 
\textbf{enhanced GA} & \textbf{93.1\%} & \textbf{95.2\%} && \textbf{79\%} & \textbf{89.4\%} \\
\midrule
\textbf{w/o Phases 1.1 \& 1.2} & 79.0\% & 82.7\% && 54.1\% & 59.1\% \\
\textbf{w/o Phase 2} & 92.6\% & 94.8\% && 78.7\% & 86.0\% \\
\textbf{w/o Phase 3} & 92.6\% & 94.8\% && 75.3\% & 82.5\% \\
\textbf{w/o mutations} & 91.2\% & 94.3\% && 73.6\% & 81.1\% \\
\midrule
\textbf{Sholomon \textit{et al.} \cite{Sholomon_2013_CVPR}} & 74.7\% & 79.1\% && 57.0\% & 62.6\%

\end{tabular}
\label{tab:ga_ablation}
\end{table}
    
To evaluate the effectiveness of our GA-based solver in achieving global optima, we compiled a set of fitness values (from the best chromosomes across all runs) and compared them with the ultimate fitness value corresponding to the ground truth for each puzzle. While the GA solver does not produce fitness values identical to the ground truth, the observed deviations were relatively minor. This indicates that our enhanced GA-based solver performs near-optimally, leveraging the compatibility scores generated by our DLCM. Specifically, the average fitness differences recorded were $0.38\%$ for Type-1 and $1.28\%$ for Type-2.  

\begin{figure*}
\centering
\begin{tabular}{c c c c c}

\textbf{scrambled} & \multicolumn{3}{c}{\textbf{$\to$ intermediate generations $\to$}} & \textbf{reconstructed} \\

\toprule

\includegraphics[width=0.17\linewidth]{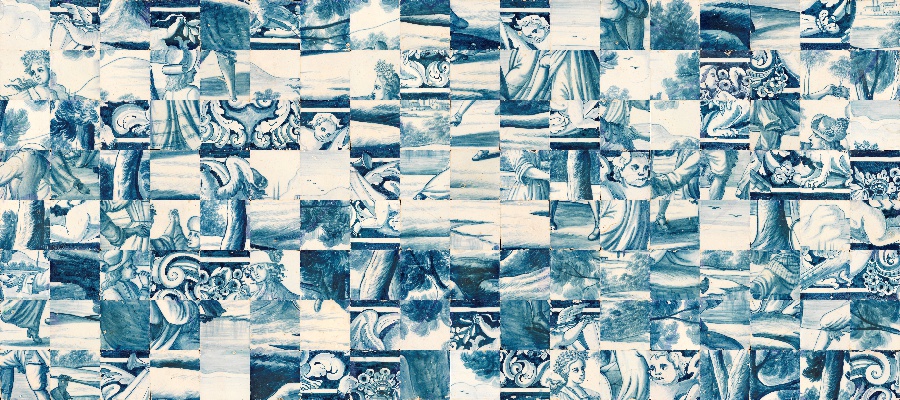} &
\includegraphics[width=0.17\linewidth]{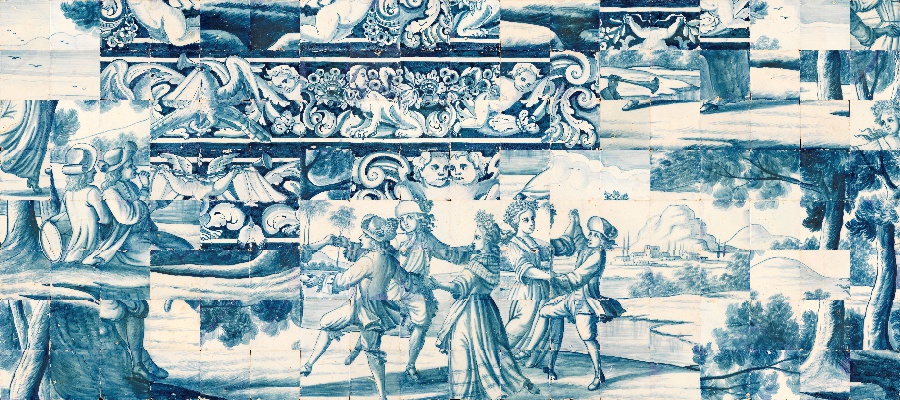} &
\includegraphics[width=0.17\linewidth]{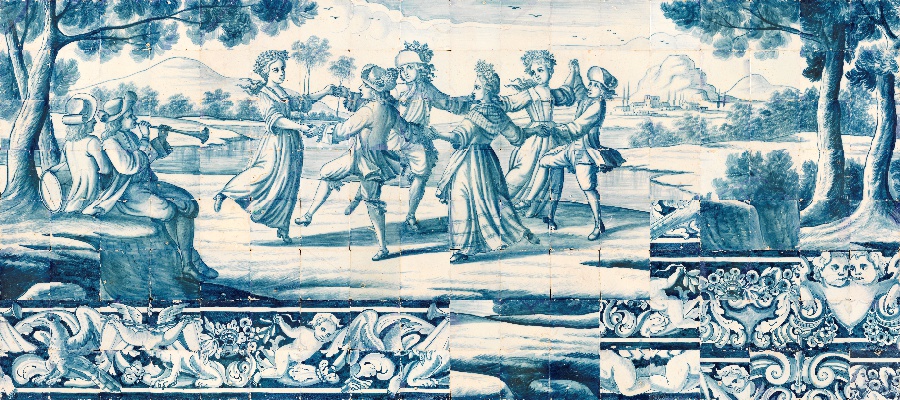} &
\includegraphics[width=0.17\linewidth]{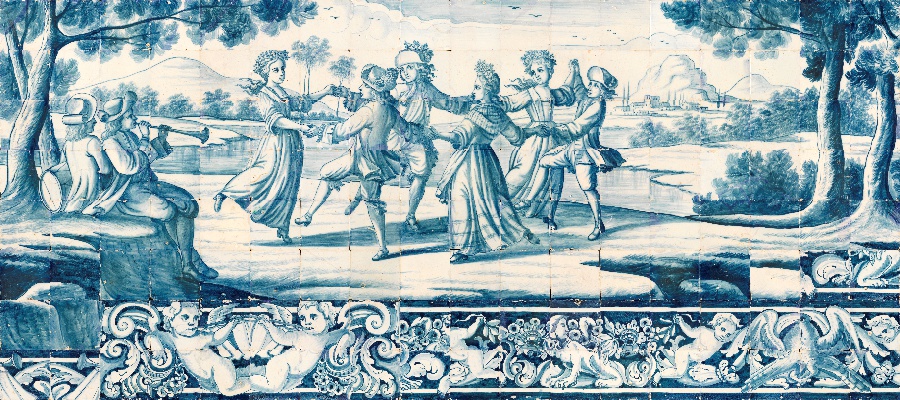} &
\includegraphics[width=0.17\linewidth]{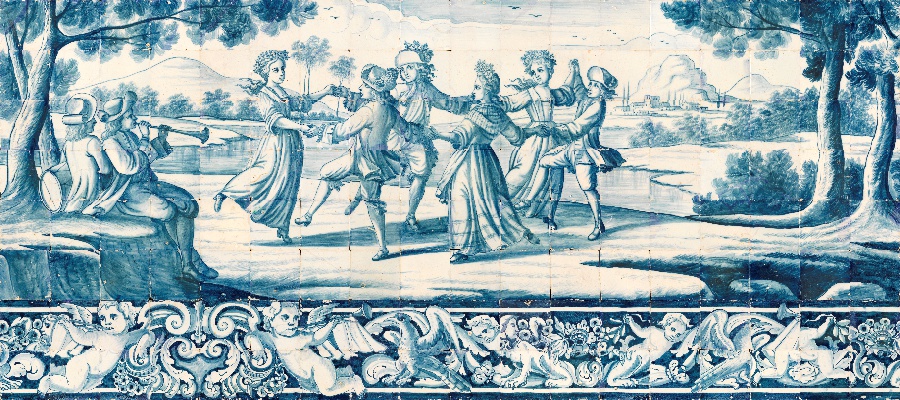}

\\

\includegraphics[width=0.17\linewidth]{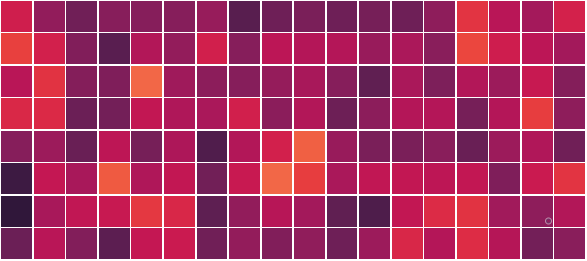} &
\includegraphics[width=0.17\linewidth]{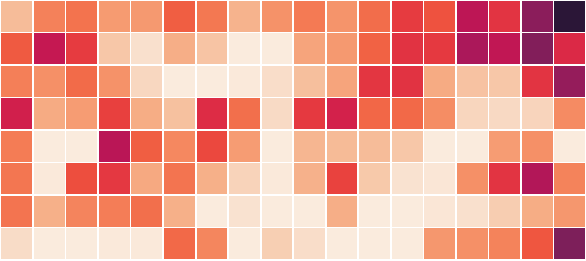} &
\includegraphics[width=0.17\linewidth]{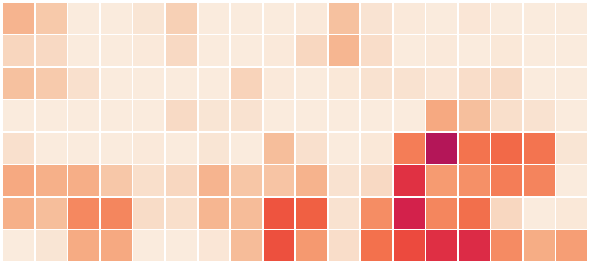} &
\includegraphics[width=0.17\linewidth]{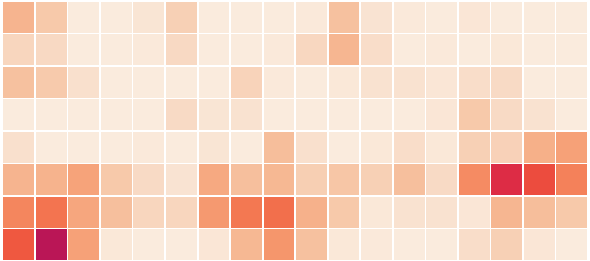} &
\includegraphics[width=0.17\linewidth]{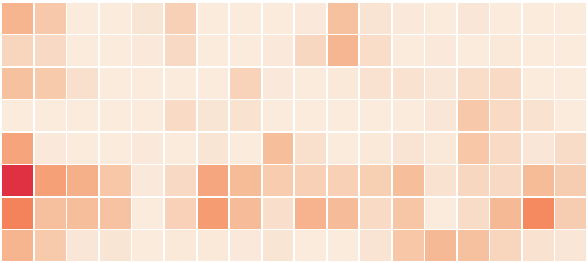}

\\

\midrule

\includegraphics[width=0.17\linewidth]{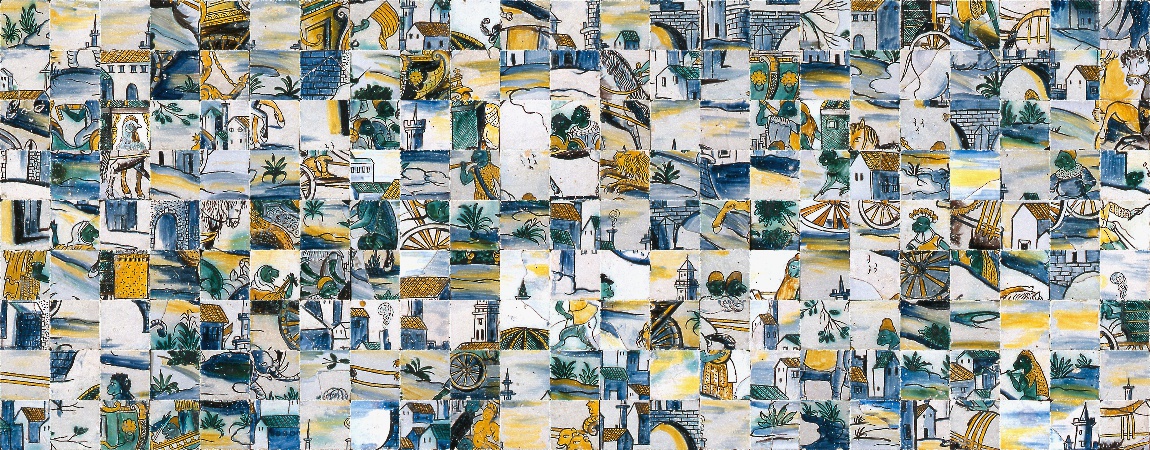} &
\includegraphics[width=0.17\linewidth]{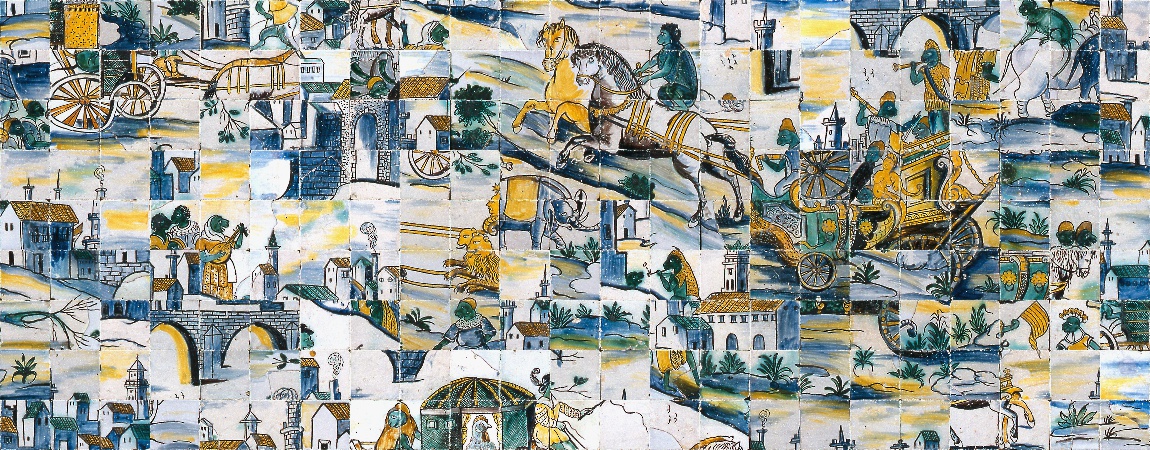} &
\includegraphics[width=0.17\linewidth]{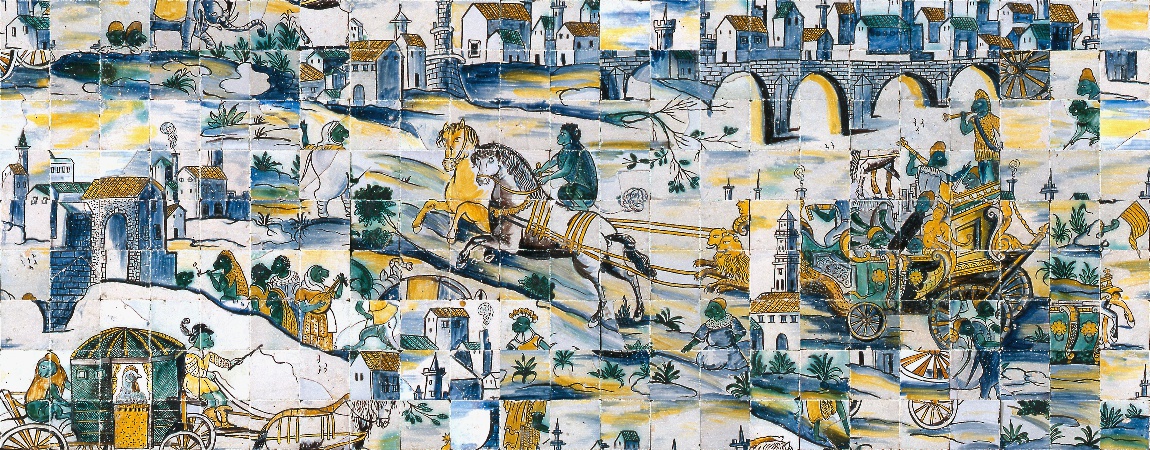} &
\includegraphics[width=0.17\linewidth]{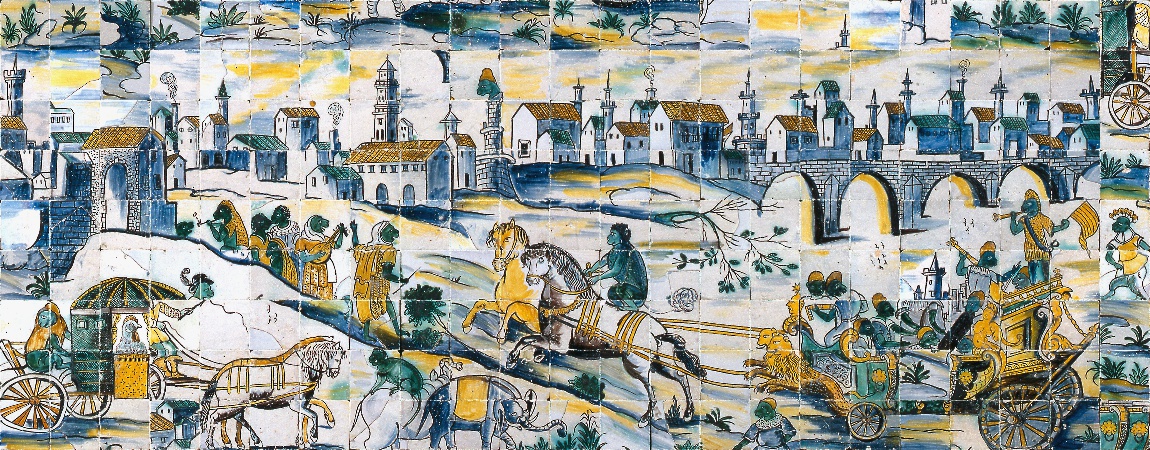} &
\includegraphics[width=0.17\linewidth]{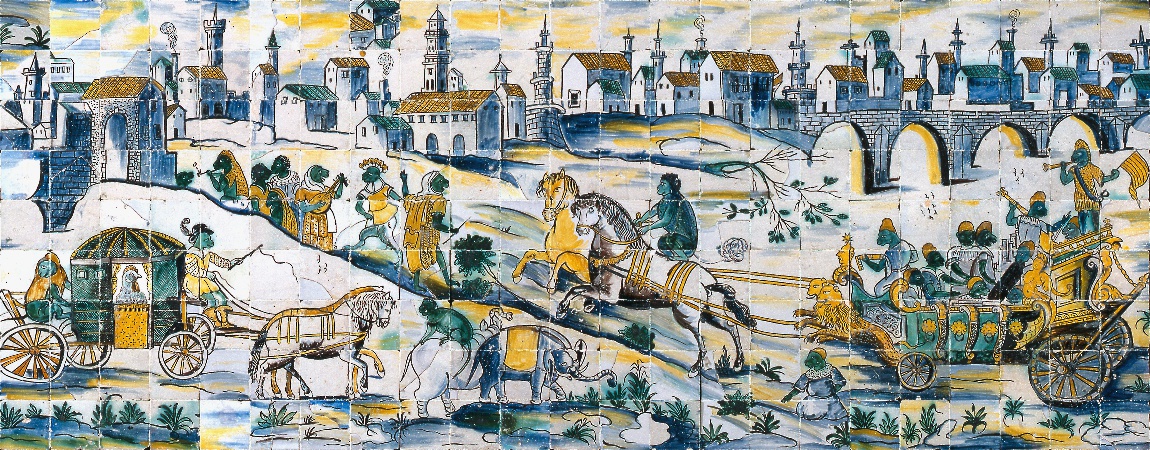}

\\

\includegraphics[width=0.17\linewidth]{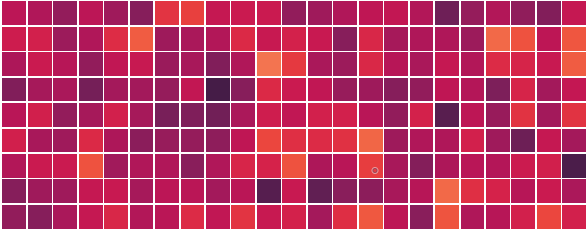} &
\includegraphics[width=0.17\linewidth]{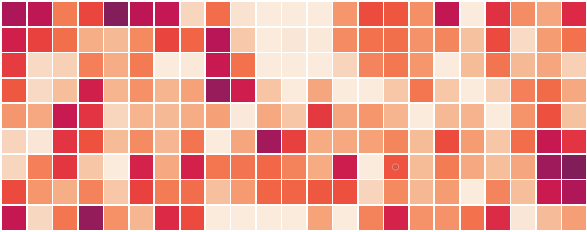} &
\includegraphics[width=0.17\linewidth]{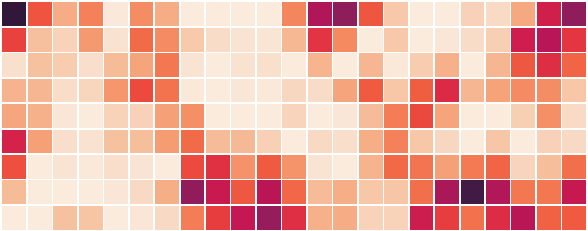} &
\includegraphics[width=0.17\linewidth]{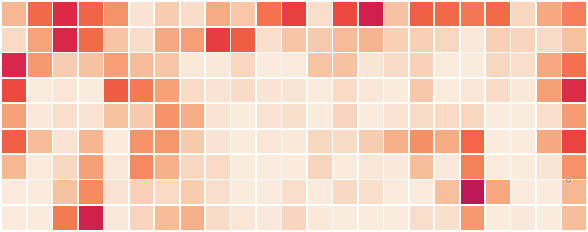} &
\includegraphics[width=0.17\linewidth]{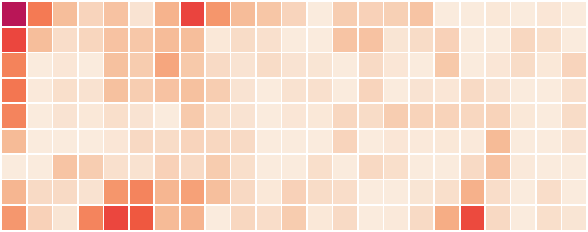}

\\

\midrule

\includegraphics[width=0.17\linewidth]{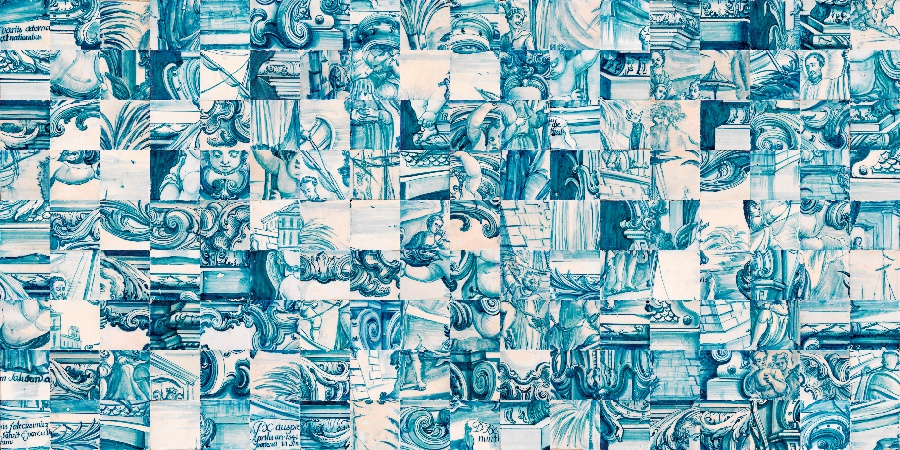} &
\includegraphics[width=0.17\linewidth]{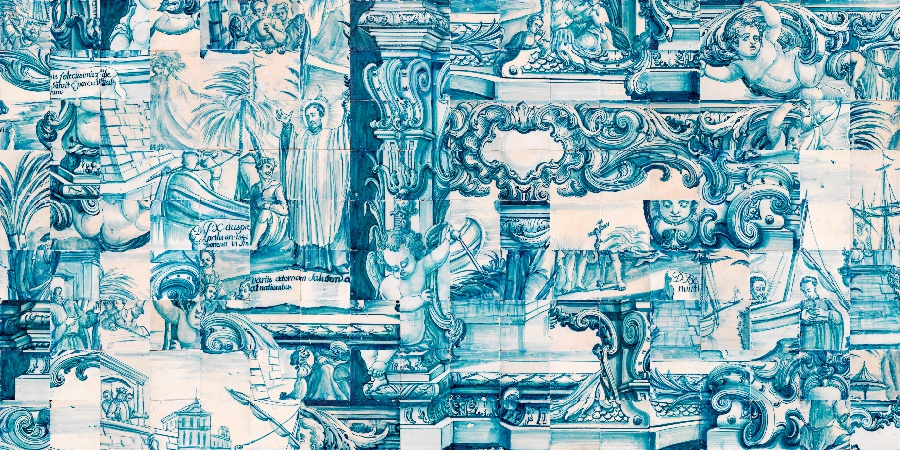} &
\includegraphics[width=0.17\linewidth]{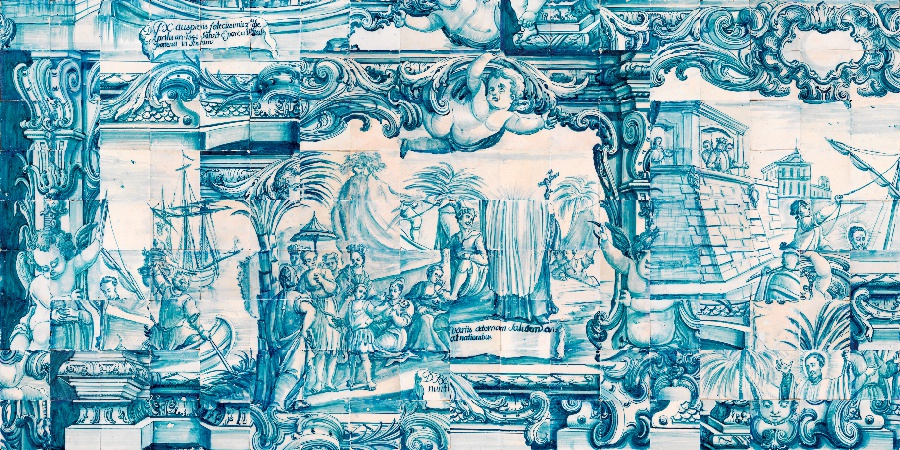} &
\includegraphics[width=0.17\linewidth]{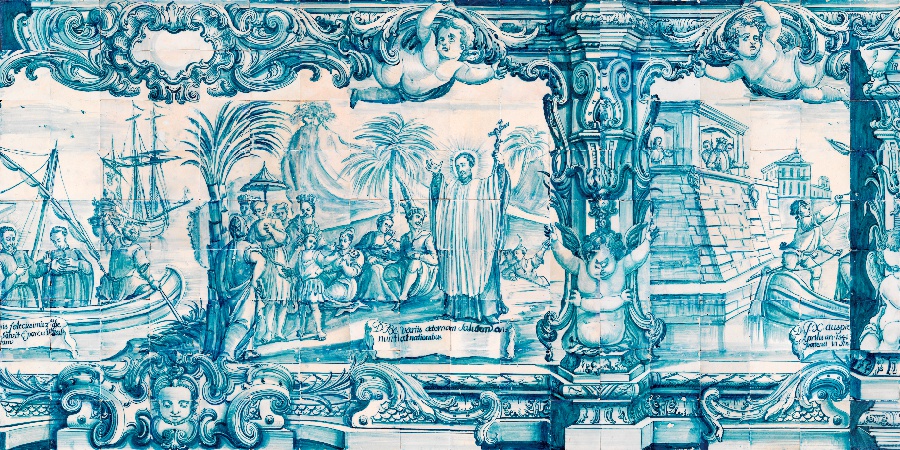} &
\includegraphics[width=0.17\linewidth]{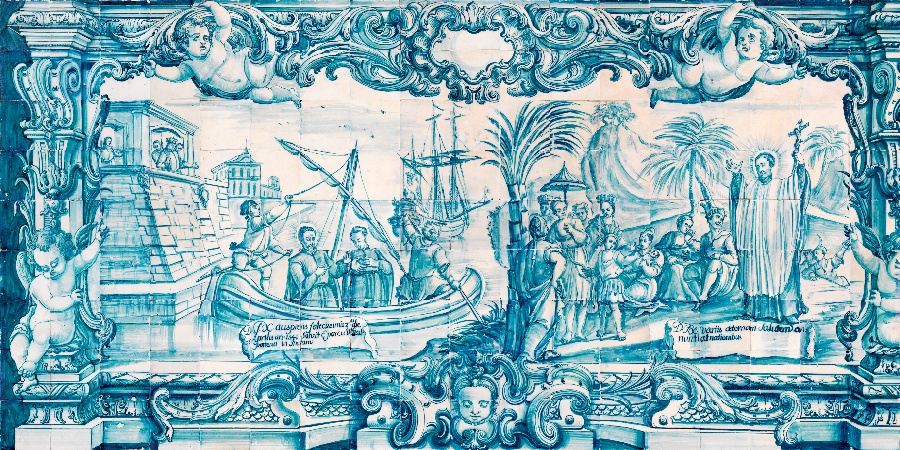}

\\

\includegraphics[width=0.17\linewidth]{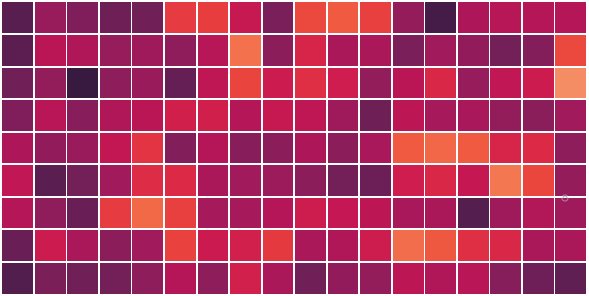} &
\includegraphics[width=0.17\linewidth]{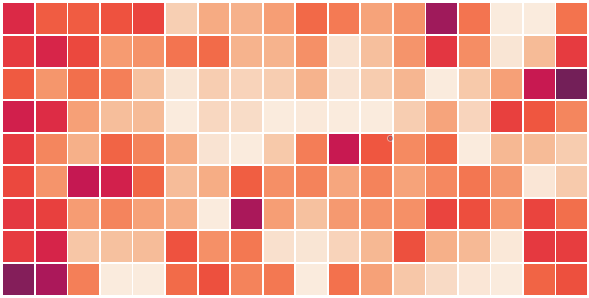} &
\includegraphics[width=0.17\linewidth]{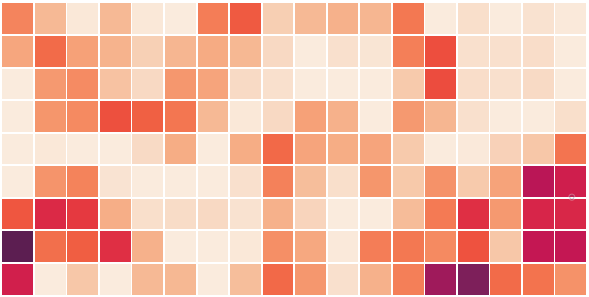} &
\includegraphics[width=0.17\linewidth]{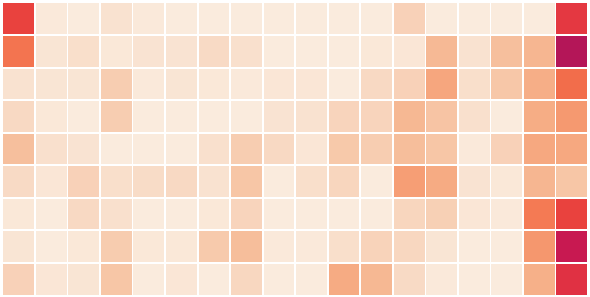} &
\includegraphics[width=0.17\linewidth]{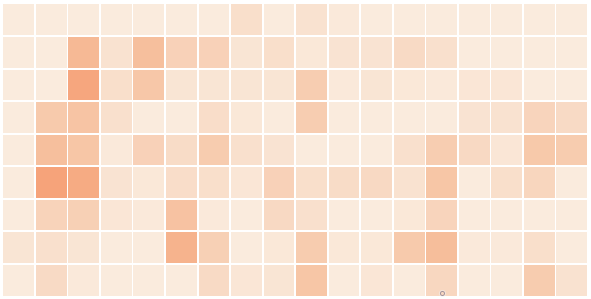}

\\

\bottomrule

\end{tabular}
\caption{Evolutionary reconstruction of three Portuguese tile panels using the enhanced GA-based solver; each sub-image shows the intermediate solution along with the corresponding "heatmap" for a specific generation, highlighting the evolutionary nature of the GA; heatmap cells represent fitness ({\ie}, average CMs) of each tile relative to its local neighborhood; the brighter a cell, the higher its local compatibility.} 
\label{fig:ga_progress}
\end{figure*}

Figure~\ref{fig:ga_progress} illustrates snapshots of the evolutionary reconstruction for three different Portuguese tile panels. These visualizations provide additional qualitative insights into the GA's progress. Each ``heatmap'' cell represents the local fitness ({\ie}, average CMs) of the puzzle piece at that position. Naturally, a successful puzzle reconstruction should correlate with high brightness in the final iteration of the heatmap.

\section{Applicability and Robustness in Additional Problem Domains}
\label{sec:other_problem_domains}
    
\begin{table*}[!t]
\centering
\caption{Neighbor accuracy of reconstruction for traditional methods compared to the proposed method on Type-1 synthetic puzzles (where $K$ represents the number of pieces); average performance of proposed method is superior in most cases.}

\begin{tabular}{|c||c|c|c|c|c|}
    \hline
    \multicolumn{6}{|c|}{\multirow{2}{*}{\textbf{{\large Type-1}}}} \\
    \multicolumn{6}{|c|}{} \\
    \hline
    \multirow{2}{*}{\textbf{Method}} & \multirow{2}{*}{\textbf{MIT (K=432)}} & \multirow{2}{*}{\textbf{McGill (K=540)}} & \multirow{2}{*}{\textbf{Pomeranz (K=805)}} & \multirow{2}{*}{\textbf{Pomeranz (K=2360)}} & \multirow{2}{*}{\textbf{Pomeranz (K=3300)}} \\
    &&&&&\\
    \hline
    \hline
    Cho {\etal}~\cite{conf/cvpr/ChoAF10} & 55\% & 0\% & - & - & - \\
    \hline
    Yang {\etal}~\cite{yang2011particle} & 86.2\% & - & - & - & - \\
    \hline
    Pomeranz {\etal}~\cite{conf/cvpr/PomeranzSB11} & 95\% & 90.9\% & 89.7\% & 84.7\% & 85\% \\
    \hline
    Andal{\'o} {\etal}~\cite{7442162} & 94.3\% & 95.3\% & 93.4\% & - & - \\
    \hline
    Gallagher~\cite{gallagher2012jigsaw} & 95.1\% & - & - & - & - \\
    \hline
    Sholomon {\etal}~\cite{Sholomon_2013_CVPR} & \textbf{96.2\%} & 96\% & 96.3\% & 88.9\% & 92.8\% \\
    \hline
    Paikin {\etal}~\cite{paikin2015solving} & 95.8\% & 96.1\% & 95.1\% & 96.3\% & 95.3\% \\
    \hline
    Yu {\etal}~\cite{yu2016bmvc} & 95.7\% & 97.3\% & 96.6\% & 97.6\% & 97.5\% \\
    \hline
    Son {\etal}~\cite{son2019tpami} & 95.6\% & 97\% & 95.5\% & 96\% & 97.7\% \\
    \hline
    Proposed & 96.1\% & \textbf{97.9\%} & \textbf{98.1\%} & \textbf{97.8\%} & \textbf{97.9\%} \\
    \hline
    \end{tabular}
\label{tab:synthetic_results_1}
\end{table*}

\begin{table}[!t]
\centering
\caption{Neighbor accuracy of reconstruction for traditional methods vs. the proposed method on synthetic Type-2 puzzles (where $K$ is the number of pieces); the proposed method achieves, on average, new SOTA across three datasets and the highest number of perfectly reconstructed images across all datasets.}
\begin{tabular}{|c||c|c|}
    \hline
    \multicolumn{3}{|c|}{\multirow{2}{*}{\textbf{{\large Type-2}}}} \\
    \multicolumn{3}{|c|}{} \\
    \hline
    \multirow{2}{*}{\textbf{Method}} & \multicolumn{2}{c|}{\textbf{MIT (K=432)}} \\
    \cline{2-3}
    & \textbf{Neighbor} & \textbf{Perfect} \\
    \hline
    \hline
    Gallagher~\cite{gallagher2012jigsaw} & 90.4\% & 9 \\
    \hline
    Sholomon {\etal}~\cite{sholomon2016dnn} & 95.7\% & 12 \\
    \hline
    Yu {\etal}~\cite{yu2016bmvc} & 95.3\% & 14 \\
    \hline
    Son {\etal}~\cite{son2019tpami} & 95.6\% & 12 \\
    \hline
    Proposed & \textbf{96\%} & \textbf{15} \\
    \hline
    \hline
    \multirow{2}{*}{\textbf{Method}} & \multicolumn{2}{c|}{\textbf{McGill (K=540)}} \\
    \cline{2-3}
    & \textbf{Neighbor} & \textbf{Perfect} \\
    \hline
    \hline
    Gallagher~\cite{gallagher2012jigsaw} & 73.3\% & 7 \\
    \hline
    Sholomon {\etal}~\cite{sholomon2016dnn} & \textbf{96.4\%} & 11 \\
    \hline
    Yu {\etal}~\cite{yu2016bmvc} & 93.3\% & - \\
    \hline
    Son {\etal}~\cite{son2019tpami} & 94.5\% & 11 \\
    \hline
    Proposed  & 96.1\% & \textbf{13} \\
    \hline
    \hline
    \multirow{2}{*}{\textbf{Method}} & \multicolumn{2}{c|}{\textbf{Pomeranz (K=805)}} \\
    \cline{2-3}
    & \textbf{Neighbor} & \textbf{Perfect} \\
    \hline
    \hline
    Gallagher~\cite{gallagher2012jigsaw} & 85.5\% & 5 \\
    \hline
    Sholomon {\etal}~\cite{sholomon2016dnn} & 95.9\% & 8 \\
    \hline
    Yu {\etal}~\cite{yu2016bmvc} & 92.9\% & - \\
    \hline
    Son {\etal}~\cite{son2019tpami} & 94.4\% & \textbf{11} \\
    \hline
    Proposed  & \textbf{96.3\%} & \textbf{11} \\
    \hline
    \hline
    \multirow{2}{*}{\textbf{Method}} & \multicolumn{2}{c|}{\textbf{Pomeranz (K=2360)}} \\
    \cline{2-3}
    & \textbf{Neighbor} & \textbf{Perfect} \\
    \hline
    \hline
    Gallagher~\cite{gallagher2012jigsaw} & 62.5\% & 0 \\
    \hline
    Yu {\etal}~\cite{yu2016bmvc} & 95.5\% & - \\
    \hline
    Son {\etal}~\cite{son2019tpami} & 96.8\% & \textbf{1} \\
    \hline
    Proposed  & \textbf{96.9\%} & \textbf{1} \\
    \hline
    \hline
    \multirow{2}{*}{\textbf{Method}} & \multicolumn{2}{c|}{\textbf{Pomeranz (K=3300)}} \\
    \cline{2-3}
    & \textbf{Neighbor} & \textbf{Perfect} \\
    \hline
    \hline
    Gallagher~\cite{gallagher2012jigsaw} & 81.9\% & \textbf{1} \\
    \hline
    Yu {\etal}~\cite{yu2016bmvc} & 90.2\% & - \\
    \hline
    Son {\etal}~\cite{son2019tpami} & \textbf{95.2\%} & \textbf{1} \\
    \hline
    Proposed  & 92.7\% & \textbf{1} \\
    \hline
    \end{tabular}
\label{tab:synthetic_results_2}
\end{table}

\begin{table}[!t]
\centering
\caption{Neighbor accuracy of reconstruction for Bridger {\etal}~\cite{bridger2020cvpr} compared to our proposed hybrid scheme on Type-1 puzzles from MIT, McGill, and Pomeranz805, with $7\%$ and $14\%$ erosion ({\ie}, 2 and 4 missing pixels along each puzzle piece boundary).}

\begin{tabular}{|c||c|c||c|c|}
    \hline
    \multicolumn{5}{|c|}{\multirow{2}{*}{\textbf{{\large 7\% Erosion}}}} \\
    \multicolumn{5}{|c|}{} \\
    \hline
    \multirow{2}{*}{\# Pieces} & \multicolumn{2}{c||}{Neighbor} & \multicolumn{2}{c|}{Perfect} \\
    \cline{2-5}
    & Bridger {\etal} & Ours & Bridger {\etal} & ours \\
    \hline
    \hline
    70 pieces & 84.6\% & \textbf{97.9\%} & 4 / 20 & \textbf{18 / 20} \\
    \hline
    88 pieces & 79.6\% & \textbf{93.2\%} & 7 / 20 & \textbf{15 / 20} \\
    \hline
    150 pieces & 76.3\% & \textbf{98\%} & 2 / 20 & \textbf{14 / 20} \\
    \hline
    \hline
    \multicolumn{5}{|c|}{\multirow{2}{*}{\textbf{{\large 14\% Erosion}}}} \\
    \multicolumn{5}{|c|}{} \\
    \hline
    \multirow{2}{*}{\# pieces} & \multicolumn{2}{c||}{Neighbor} & \multicolumn{2}{c|}{Perfect} \\
    \cline{2-5}
    & Bridger {\etal} & Ours & Bridger {\etal} & ours \\
    \hline
    \hline
    70 pieces & 57.1\% & \textbf{92.2\%} & 1 / 20 & \textbf{7 / 20} \\
    \hline
    88 pieces & 51.1\% & \textbf{85.4\%} & 0 / 20 & \textbf{10 / 20} \\
    \hline
    150 pieces & 51.3\% & \textbf{87.1\%} & 0 / 20 & \textbf{5 / 20} \\
    \hline
    \end{tabular}
\label{tab:eroded_boundaries_comparison_7}
\end{table}
    
We further investigate the performance of our scheme to explore its broader applicability beyond the Portuguese tile problem domain. Specifically, we extend the analysis to the following problem domains: (1) Synthetic JPP, (2) eroded boundaries, and (3) shredded documents. Our results show that our method achieves state-of-the-art performance in the first two domains and demonstrates promising potential on the \textit{strip-cut} variant of the shredded documents problem. The results strongly support the extended applicability and robustness of our hybrid methodology across various real-world reconstruction tasks, framed as puzzle reconstruction of non-overlapping square pieces.

\subsection{Synthetic JPP}
This classical version of a realistic image decomposed into non-overlapping ($28 \times 28$) squares is one of the most frequently studied cases of the JPP. Several benchmarks have been established to evaluate the performance of various puzzle reconstruction algorithms developed over the years. These benchmarks include: (1) The MIT dataset~\cite{conf/cvpr/ChoAF10}, which consists of 20 432-piece images, (2) the McGill dataset~\cite{mcgill_dataset}, which includes 20 540-piece images, and (3) the datasets compiled by Pomeranz et al.~\cite{conf/cvpr/PomeranzSB11}, which contain 20 805-piece images, three 2360-piece images, and three 3300-piece images.
We also tested our proposed scheme on these datasets, which have become a standard testbed for evaluating most puzzle reconstruction algorithms.

To train our DLCM module for this domain, we used the DIV2K dataset~\cite{Agustsson_2017_CVPR_Workshops}, consisting of 800 training images and 100 additional validation images. We then employed our GA-based solver on all images for both Type-1 and Type-2 variants. The results are presented in Tables~\ref{tab:synthetic_results_1} and~\ref{tab:synthetic_results_2}, where the average best results obtained over five runs of our scheme per image are shown. Our method consistently achieves higher reconstruction accuracy than all other methods tested on nearly every dataset considered.

\subsection{Eroded Boundaries}
\begin{figure}[!t]
    \centering
    \subfloat[Bridger {\etal}]{\includegraphics[width=0.45\columnwidth]{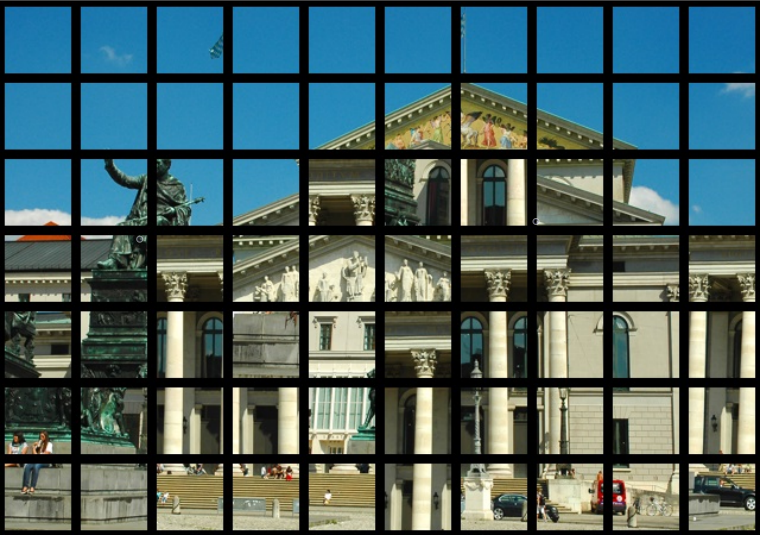}}
    \hfill
    \subfloat[Ours]{\includegraphics[width=0.45\columnwidth]{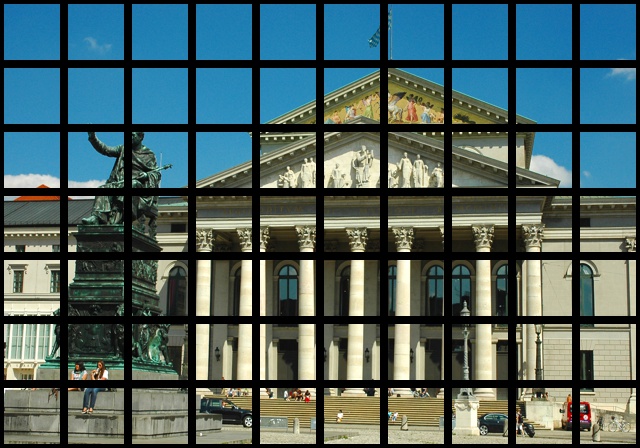}}
    \\
    \subfloat[Bridger {\etal}]{\includegraphics[width=0.45\columnwidth]{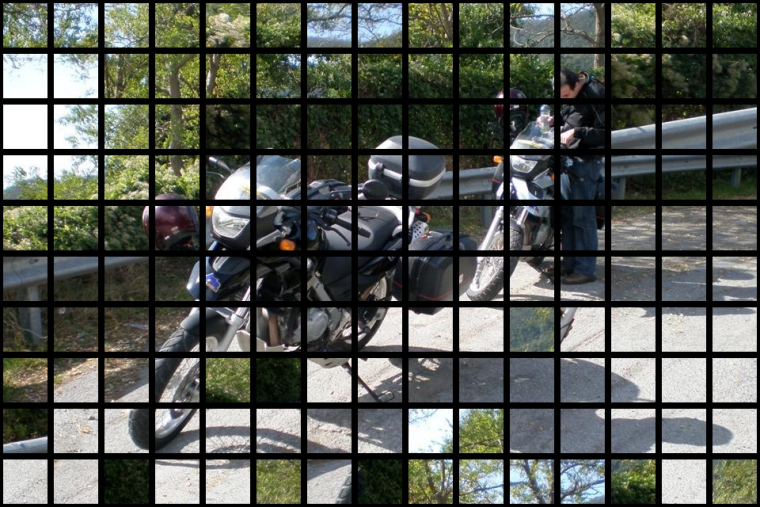}}
    \hfill
    \subfloat[Ours]{\includegraphics[width=0.45\columnwidth]{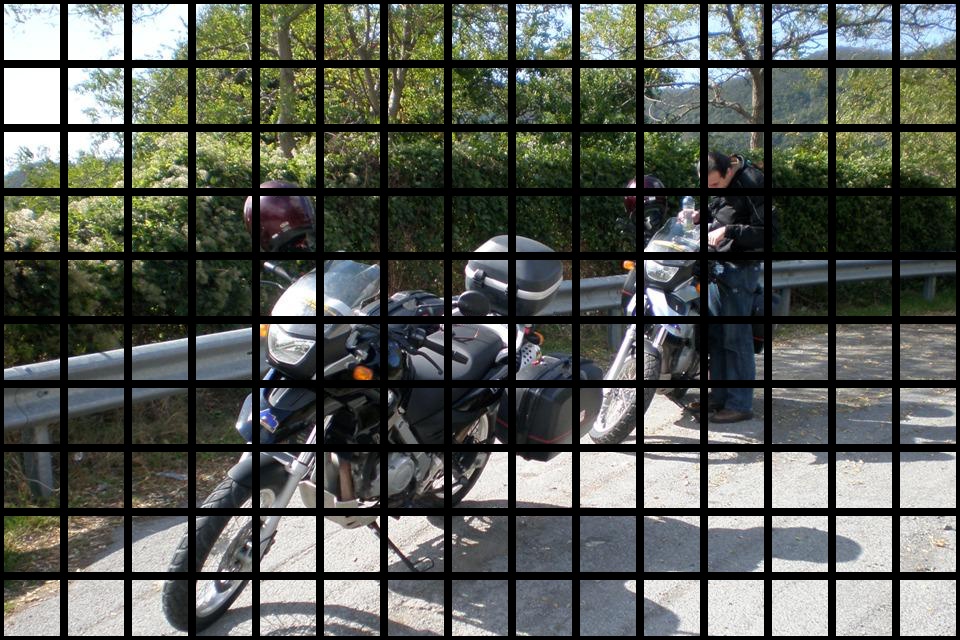}}
    \caption{Reconstruction of 70- and 150-piece puzzles with 14\% erosion: (a), (c) assembled using~\cite{bridger2020cvpr}, and (b), (d) obtained with our proposed method, which better reflects the global context of the scene.}
    \label{fig:eroded_boundaries_illustration}
\end{figure}

Mondal {\etal}~\cite{mondal2013crv} first tackled the challenge of eroded boundaries for 7\% and 14\% erosion, corresponding to $t=1$ and $t=2$. They used $28 \times 28$ pieces, the same piece size as in the synthetic JPP. Bridger {\etal}~\cite{bridger2020cvpr} later proposed their GAN-based approach. Unlike~\cite{mondal2013crv}, their method works with $64 \times 64$ pieces to provide more contextual information to the \textit{generator}, yielding more precise in-painting while retaining the $7\%$ and $14\%$ erosions. Their method was trained on the DIV2K dataset~\cite{Agustsson_2017_CVPR_Workshops} and evaluated on the MIT, McGill, and Pomeranz805 datasets. Due to the larger pieces, the transformed datasets contain puzzles with 70-, 88-, and 150-piece puzzles (instead of 432-, 540-, and 805-piece puzzles, respectively). For a fair comparison, we trained our DLCM with the same piece size on the DIV2K dataset~\cite{Agustsson_2017_CVPR_Workshops}, applying a random degree of erosion between $7\%$ and $14\%$, {\ie}, 2 to 4 missing pixels. After applying our GA-based solver for reconstruction, we achieved a significant improvement over Bridger {\etal}~\cite{bridger2020cvpr}. Specifically, our method improved their previous SOTA accuracy by 16.2\% and 35.1\% on average for 7\% and 14\% erosion, respectively. See Table~\ref{tab:eroded_boundaries_comparison_7} and Figure~\ref{fig:eroded_boundaries_illustration} for a detailed comparison.

Most recently, Liu \textit{et al.} reported partial results on the same dataset (see \cite{liu2024diffusion}, Table 3). Their method yielded 75.9\% direct accuracy and 45\% perfect reconstruction on Type-1 150-piece puzzles with 7\% erosion. In contrast, our scheme achieved superior 98\% neighbor accuracy and 70\% perfect reconstruction.

\begin{figure*}[h]
\centering
    \subfloat[Scrambled]{\includegraphics[width=0.9\linewidth]{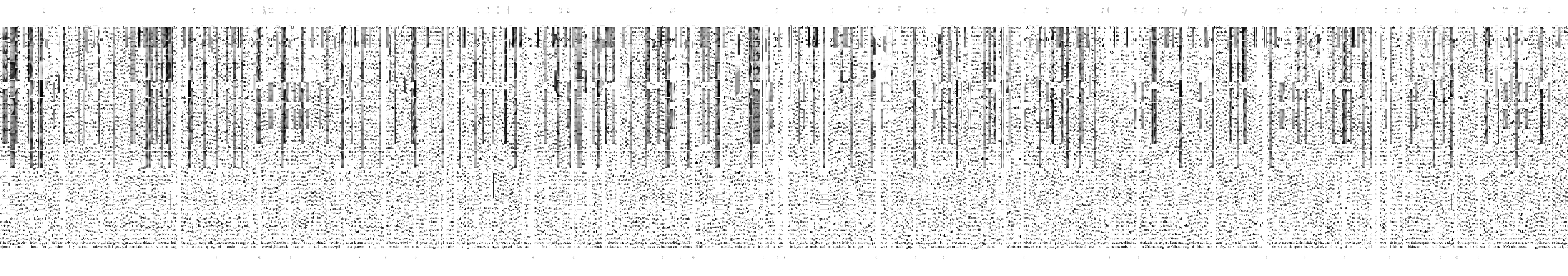}}
    \\
    \subfloat[Reconstructed]{\includegraphics[width=0.9\linewidth]{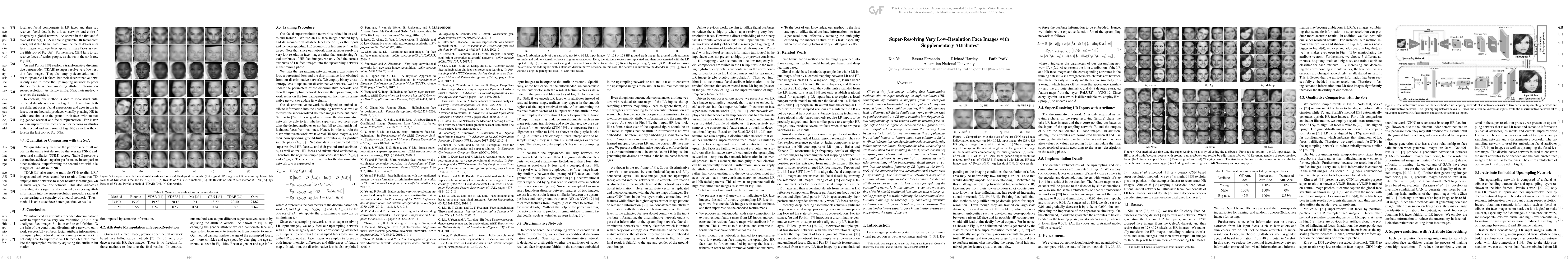}}
    \caption{Reconstruction of a puzzle made up of nine shredded documents: (a) Scrambled puzzle with 765 strips of width 2.47mm, and (b) readable reconstruction with $97.1\%$ neighbor accuracy achieved by our proposed method.}
\label{fig:shredded_big_puzzle}
\end{figure*}

\subsection{Shredded Documents}
The reconstruction of \textit{shredded documents} has garnered significant attention in recent years. In particular, the commonly addressed \textit{strip-cut} problem remains a challenging task. Shredded documents typically suffer from a severe loss of information, consisting mainly of sparse black segments on a white background. Numerous studies have been conducted in this domain using real-world datasets, which often require extensive alignment preprocessing of strip pairs.

In contrast, we utilized a self-constructed dataset to test our scheme, aiming to demonstrate its effectiveness as part of a more complex system. Specifically, we created our own synthetic strip-cut dataset using all the CVPR '18 papers. The dataset contains 9,474 grayscale pages, divided into 80\% for training, 10\% for validation, and 10\% for testing.

\textbf{GrayNet}. As mentioned earlier, our DLCM generally consists of four sub-networks: three that handle each color channel separately, and a fourth that processes all three RGB channels. However, in the case of shredded documents, which are typically grayscale images, the input consists of only a single channel. Therefore, for this problem, our DLCM model is adapted to a single sub-network called GrayNet, which is four times smaller than the regular DLCM scheme. The training process is essentially the same as before, but here we used $50 \times 50$ pieces, i.e., an input pair of size $50 \times 100 \times 1$. The piece size was determined based on the number of strips per A4 page, and the final layer of our network architecture was adjusted accordingly.

\begin{figure}[h]
    \centering
    \includegraphics[width=0.8\columnwidth]{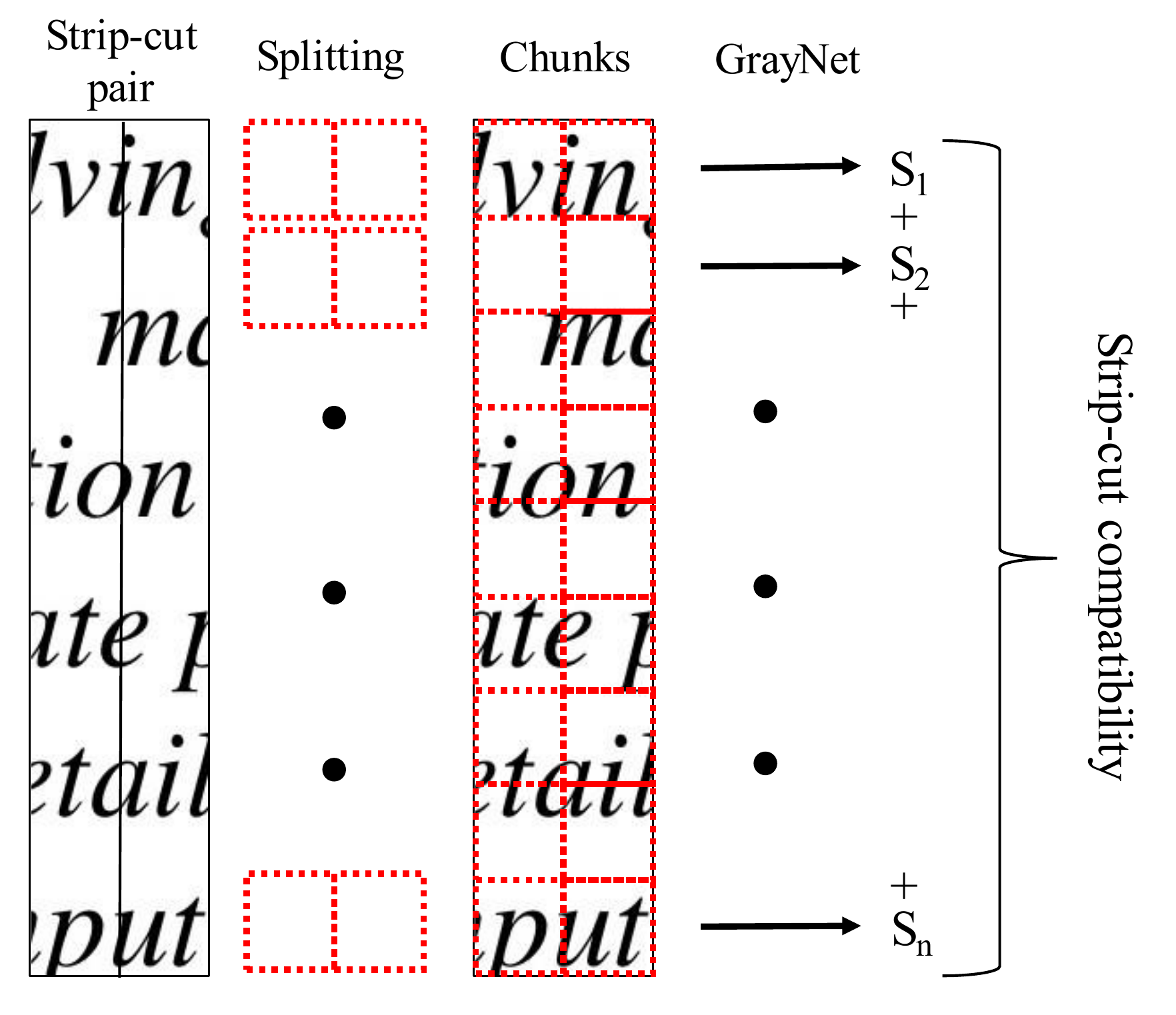}
    \caption{Compatibility measure for strip-cut pair. Left to right: Two strips, adjusted as a strip-cut pair, are divided into $n$ chunks (with any residue disposed of in subsequent phases); GrayNet assigns a compatibility score $S_i$ to each chunk $X_i$, and the final strip-cut compatibility is determined by the sum of the scores.}
    \label{fig:strip_cut_score}
\end{figure}

\textbf{Score of a strip-cut pair}. To further tailor our DLCM module to the characteristics of strip-cut documents, where strip height is typically much greater than its width, we split each strip into chunks the same size as the input to GrayNet. Each chunk $X_i$ is then assigned a score $S_i$ by GrayNet, and the total score $S$ of a strip-cut pair is the sum of all chunk scores, i.e., $S = \sum_{i=1}^{n}{S_i}$. Figure~\ref{fig:strip_cut_score} illustrates this evaluation process. (We disregard any remaining pixels beyond the largest integer multiple of 50 pixels along each strip.)

\textbf{Results}. We tested our method on a 9-page CVPR '18 paper with a resolution of 500 dpi (resulting in an image size of $5500 \times 4250$ pixels). The page size was chosen to yield 85 strips of size $5500 \times 50$ (i.e., 2.47mm width) per page. We first attempted to reconstruct each of the nine pages individually, achieving an average reconstruction accuracy of 99.5\%. To increase the difficulty, we created a second puzzle from the 765 strips of all nine pages. Our method successfully reconstructed this larger puzzle with 97.1\% accuracy, as shown in Figure~\ref{fig:shredded_big_puzzle}. These results were sufficient to recover all the relevant information from the examined paper.

\section{Conclusion}\label{sec:conclusions}
This paper introduces a generic hybrid framework for addressing 2D real-world reconstruction problems, formulated as a Jigsaw Puzzle Problem (JPP) with square, non-overlapping pieces. Our approach integrates an innovative deep learning-based compatibility measure (DLCM) model with an enhanced genetic algorithm (GA)-based solver. The DLCM model assesses the compatibility of puzzle piece pairs by analyzing their \textit{entire} content rather than relying solely on adjacent boundaries, aiming to create a robust compatibility measure for handling degraded puzzles common in real-world scenarios. The reconstruction process then employs the enhanced GA-based solver to achieve global optimization based on the pairwise DLCM scores, resulting in consistently robust performance across diverse datasets and problem domains.

Empirical results establish the state-of-the-art (SOTA) performance of our framework on large Type-1 and Type-2 puzzles, including synthetic JPP, Portuguese tile panels, and degraded puzzles with eroded boundaries. Additionally, the framework demonstrates strong potential for reconstructing strip-cut shredded documents.

To further adapt this framework to practical real-world challenges, we aim to tackle issues such as missing pieces, unknown puzzle dimensions, and multiple simultaneous puzzles. We also plan to employ computational innovations based on embeddings~\cite{twinembedding2022, edge2vec2022} to alleviate significantly  the current computational bottleneck of the CM model, which involves calculating $16N^2$ pairwise compatibility scores (where $N$ is the number of pieces). Reducing the computational burden of compatibility measure calculations will enable the development of faster and more efficient DL-based reconstruction models. Regarding the running times of our enhanced GA solver, we intend to improve the relatively slow running times reported in~\cite{Sholomon_2013_CVPR, sholomon2014genetic, sholomon2014generalized} through the implementation, for example, of a multi-threaded solution.

\section*{Acknowledgments}
We sincerely thank the National Museum of the Azulejo (MNAz) in Lisbon, Portugal, for their invaluable collaboration and for providing datasets that have been instrumental in advancing our research on the Portuguese tile problem.

\appendices
\section{Compatibility Map}
\label{sec:compatibility map}
\begin{figure}
    \centering
    \tabular{c}
    \includegraphics[width=0.70\columnwidth]{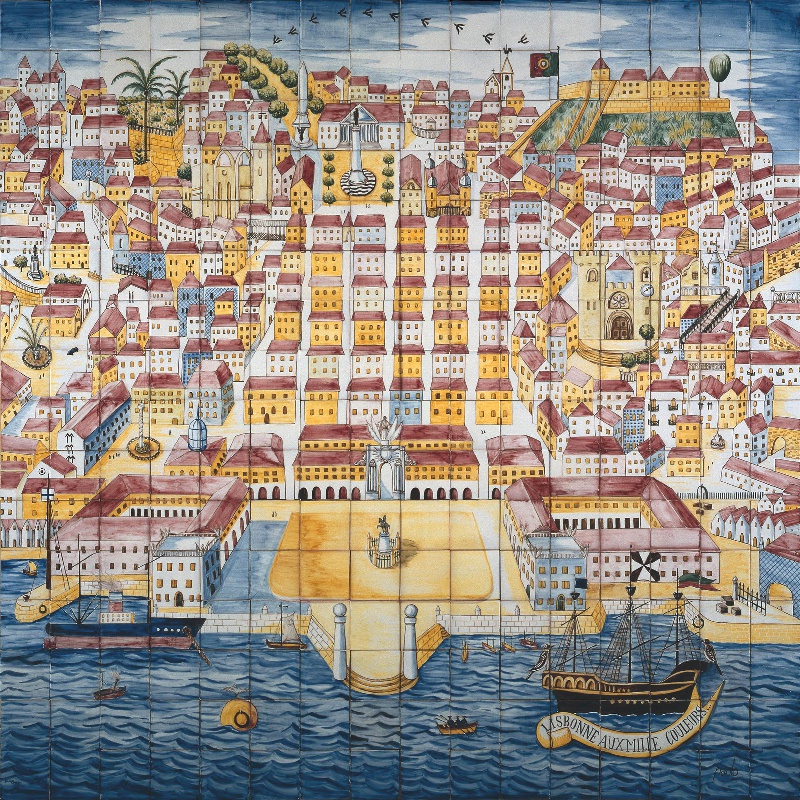} \\
    \textbf{Portuguese Panel} \\
    \includegraphics[width=0.70\columnwidth]{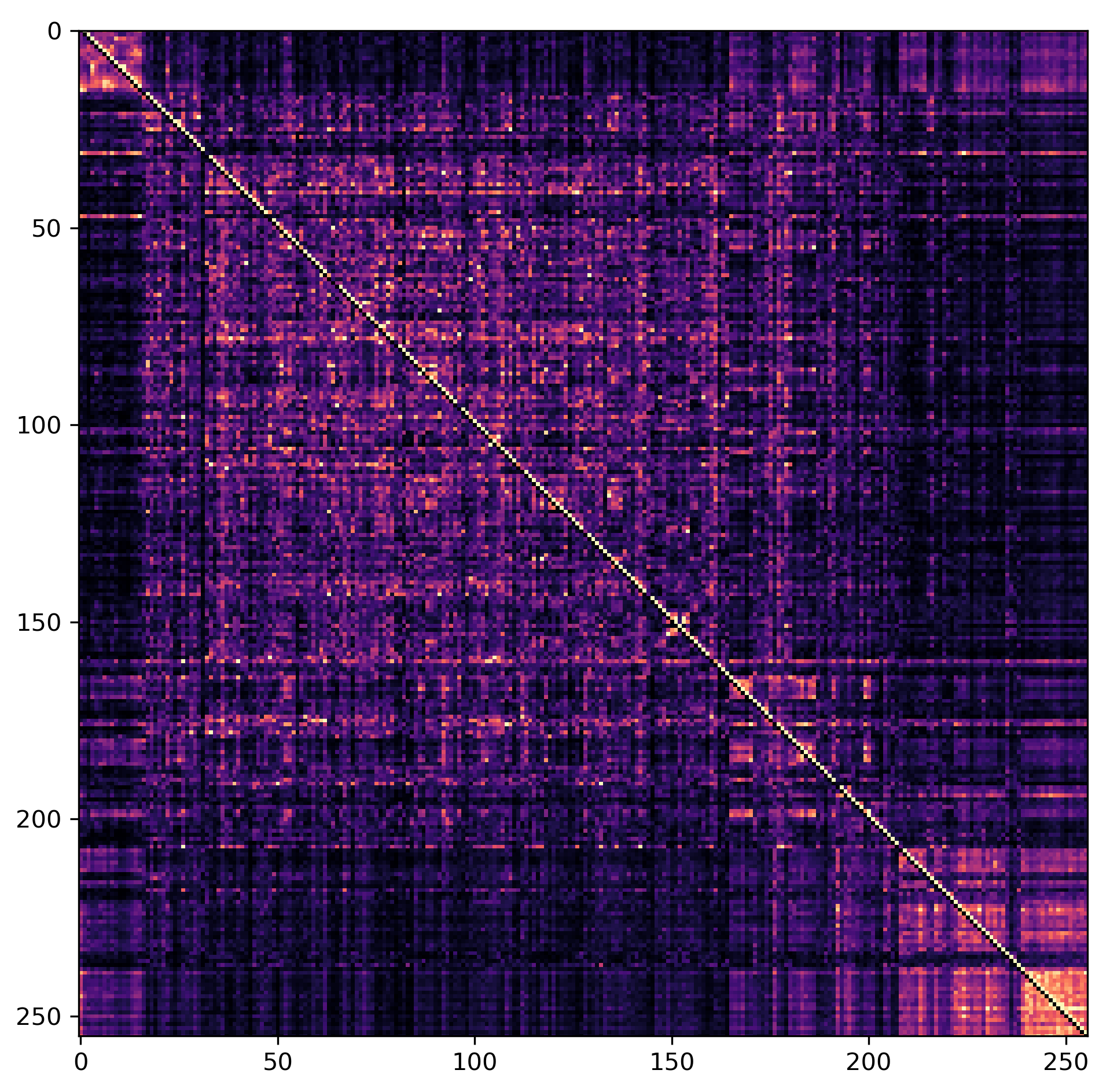} \\
    \textbf{DLCM Score Map}
    \endtabular
    
    \caption{Compatibility map of puzzle tiles: (Top) 256-piece Portuguese tile panel from the MNAz test set, and (bottom) compatibility score matrix; the $i$-th row represents anchor piece $i$, and the $j$-th column represents examined piece $j$ with respect to the right edge of the anchor; unless the anchor is positioned at the rightmost puzzle, its correct neighboring piece on the right is piece $(i+1)$, {\ie}, super diagonal entries of the CM matrix should be the highest (in yellow), while other entries (in dark purple) should be lower.}
    \label{fig:dlcm_score_map}
\end{figure}

The performance of our DLCM is effectively illustrated through a compatibility score map (or matrix) for a given Portuguese tile panel, as shown in Figure~\ref{fig:dlcm_score_map} (representing one of the 16 possible matrices). The bottom figure presents the compatibility score matrix generated by our DLCM for the 256-piece Portuguese tile panel from the MNAz test set (depicted at the top). 

In this matrix, the $(i, j)$-th entry corresponds to the compatibility score between the anchor piece $i$ and the candidate piece $j$, specifically for the right edge of the anchor. If the anchor is not the rightmost tile of a row, the correct neighboring piece to its right is piece $i+1$, assuming the tiles are sequentially numbered from top to bottom and left to right. As a result, the highest compatibility scores (highlighted in yellow) should appear along the super-diagonal of the CM matrix, while other scores, expected to be significantly lower, are represented in dark purple. 

This stark contrast between the bright super-diagonal entries and the dark regions elsewhere in the matrix offers a qualitative assessment tool for evaluating the effectiveness of our DLCM scheme.

\section{Other DL-Based CMs}
\label{sec:other_DL_CMs}
We describe below the retraining of the SqueezeNet-based V1.1 model~\cite{paixao2018deep} and the GAN-based model~\cite{bridger2020cvpr} on our Portuguese tile dataset, as these DL-based models could potentially improve accuracy and robustness by leveraging full content information. Both of these DL-based CMs were retrained on the same training and validation sets to ensure a fair comparison with our DLCM.

\textbf{SqueezeNet}:
\newline
We retrained the pre-trained SqueezeNet V1.1 model on the $50 \times 50$ Portuguese tile dataset for comparison with our DLCM. The same data augmentations used for our DLCM were applied. Apart from this, all training hyper-parameters reported in~\cite{paixao2018deep} were left unchanged.

\textbf{GAN-based}:
\newline
Bridger {\etal}~\cite{bridger2020cvpr} utilized $64 \times 64$-tile puzzles, with 7\% and 14\% of the pixels near the boundaries removed to simulate erosion (see Figure~\ref{fig:eroded_boundaries} for an example). To adapt their GAN-based CM for the Portuguese tile problem, we resized all of our training and test images to $64 \times 64$ pixels, instead of the $50 \times 50$ size used by our DLCM, ensuring compatibility with their GAN architecture, which requires an input size that is a power of 2.

\begin{figure}[!t]
    \centering
    \subfloat[]{\includegraphics[width=0.5\columnwidth]{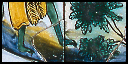}}
    \hfil
    \subfloat[]{\includegraphics[width=0.5\columnwidth]{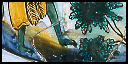}}
    \\
    \subfloat[]{\includegraphics[width=0.5\columnwidth]{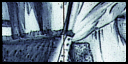}}
    \hfil
    \subfloat[]{\includegraphics[width=0.5\columnwidth]{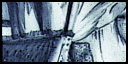}}
    \caption{Illustration of in-painting results generated by the \textit{generator} of~\cite{bridger2020cvpr} on two piece pairs from our Portuguese tile test set: (a), (c) Original pairs, and (b), (d) corresponding in-painting results.}
    \label{fig:in_painting_pairs}
\end{figure}

The first training phase, which was performed on eroded DIV2K data~\cite{Agustsson_2017_CVPR_Workshops}, was retained without changes, as the Portuguese tile dataset does not include in-painting ground truth required for training. However, during Phase-2, the pretrained discriminator was fine-tuned on in-painted Portuguese tile panels to specialize exclusively in this task. See Figure~\ref{fig:in_painting_pairs} for examples of in-painting results for a pair of Portuguese tiles. All other hyper-parameters in~\cite{bridger2020cvpr} were kept intact.

\bibliographystyle{IEEEtran}
\bibliography{egbib}

\end{document}